\documentclass[11pt, letterpaper, logo]{googledeepmind}

\PassOptionsToPackage{comma,numbers,sort,compress}{natbib}
\usepackage[comma,numbers,sort,compress]{natbib}
\hypersetup{
  colorlinks = true,
  linkcolor  = {red!70!black},
  citecolor  = {magenta},
  urlcolor   = {blue!80!black},
}

\pdftrailerid{redacted}

\makeatletter
\renewcommand\bibentry[1]{\nocite{#1}{\frenchspacing\@nameuse{BR@r@#1\@extra@b@citeb}}}
\makeatother

\usepackage{graphicx}
\usepackage{wrapfig}
\usepackage{fontawesome5}
\usepackage{gdm-colors}
\usepackage[noend]{algpseudocode}

\usepackage{ragged2e}
\usepackage{marvosym}
\newenvironment{paperquote}
  {\begin{quote}\itshape\justifying}
  {\end{quote}}
\usepackage{mathtools}
\usepackage{bm}
\usepackage{algorithm}      % float wrapper; pseudocode via algpseudocode in main.tex
\usepackage{multirow}
\usepackage{adjustbox}
\usepackage{verbatim}   % provides the `comment' environment

\theoremstyle{plain}

\theoremstyle{definition}

\theoremstyle{remark}

\usepackage{float}
\usepackage{placeins}   % \FloatBarrier keeps floats inside their section
\usepackage{setspace}
\usepackage{rotating}   % sideways appendix figures
\definecolor{spblue}{HTML}{00b5ea}
\definecolor{spred}{HTML}{E74C3C}
\definecolor{spgreen}{HTML}{2ECC71}
\definecolor{lightblue}{RGB}{173,216,230}
\definecolor{superlightred}{RGB}{255,204,203}
\definecolor{lightpink}{rgb}{0.98, 0.88, 0.95}
\definecolor{softblue}{RGB}{38, 139, 210}
\definecolor{softgreen}{RGB}{0, 153, 0}

\usepackage{array}
\newcommand{\std}[1]{\scriptsize{$\pm$#1}}

\usepackage{amsmath,amsfonts,bm}

\def\eqref#1{equation~\ref{#1}}
\def\1{\bm{1}}

\DeclareMathAlphabet{\mathsfit}{\encodingdefault}{\sfdefault}{m}{sl}
\SetMathAlphabet{\mathsfit}{bold}{\encodingdefault}{\sfdefault}{bx}{n}

\graphicspath{{figures/}}
\DeclareRobustCommand{\mail}[1]{\href{mailto:#1}{\Letter}}

\newcounter{takeawaycounter}

\newcommand{\takeaway}[1]{%
  \par\noindent%
  \begin{minipage}{\linewidth}%
    \refstepcounter{takeawaycounter}%
    {\setlength{\fboxsep}{2pt}%
     \colorbox{gray!20}{\textbf{\ding{224}~Observation \thetakeawaycounter.}}}~\textit{#1}%
  \end{minipage}\par\vspace{0.5em}%
}

\usepackage{tcolorbox}

\title{Stream of Belief: Eliciting Collective Intelligence through Latent Belief Dynamics}

\correspondingauthor{Correspondence to yu\_yao@mit.edu, ydu@seas.harvard.edu}

\author[1, \mail{yu_yao@mit.edu}]{Yu Yao}
\author[3, \mail{jiayidong@ucla.edu}]{Jiayi Dong}
\author[2, \mail{ydu@seas.harvard.edu}]{Yilun Du}

\affil[1]{MIT}
\affil[2]{Harvard}
\affil[3]{UCLA}

\begin{abstract}
\vspace{-18pt}
{{\ttfamily
\href{https://github.com/AI4SciDisc/stream-of-belief}
{\textcolor{black}{\faGithub} \footnotesize{\textcolor{magenta}{Embodied-Minds-Lab/Stream-of-Belief}}}
\\[0.6em]
}}
In this work, we view multi-agent reasoning as a process of seeking a collective equilibrium of implicit belief dynamics and introduce \textit{Stream of Belief} (SoB), where agents maintain continuous latent beliefs and iteratively refine them through communication and reflection. Our method seeks a collective belief state of a multi-agent system where further communication induces no meaningful information propagation. Through extensive experiments, we demonstrate the generalization of SoB across tasks and model families. More Broadly, our results suggest a new perspective on scaling multi-agent reasoning from belief dynamics.
\end{abstract}

\begin{document}

\maketitle

\section{Introduction}
\label{sec:introduction}

Collaboration among language agents is becoming an important paradigm for collective intelligence in the era of Agentic AI \citep{acharya2025agentic, guo2024large, tran2025multi}. Recent advances in multi-agent systems (MAS) \citep{li2023camel, wu2024autogen, hong2024metagpt, hu2025owl} have extended reasoning from individual models to collaborative systems of interacting agents, enabling applications in mathematical and scientific reasoning \citep{du2024debate, zhou2025reso, swanson2025virtual, gottweis2026accelerating, Eywa2026}, automated code generation \citep{qian2024chatdev, islam2024mapcoder, tao2024magis}, and open-domain question answering \citep{chen2024reconcile, zhang2024chain, li2025flow}.

This paradigm echoes a similar pattern in human societies: collective intelligence. In human societies, individuals with incomplete knowledge interact with each other and continually revise their beliefs through interaction, enabling collective capabilities that exceed those of isolated individuals \citep{galton1907vox, surowiecki2004wisdom, hong2004groups, woolley2010evidence, couzin2009collective}. Classical studies of social influence and opinion dynamics model how individuals revise their beliefs in response to others, revealing how local interactions shape the evolution of collective states \citep{degroot1974reaching, friedkin1990social, hegselmann2002opinion, deffuant2000mixing, castellano2009statistical}. Subsequent work on social learning further shows that the structure of the interactions determines the final conclusions \citep{golub2010naive, jadbabaie2012nonbayesian, lorenz2011social, becker2017network}. These studies motivate us to rethink collective reasoning through how individual beliefs are represented and updated through interaction.

However, translating this belief dynamics perspective into language-model agents raises two fundamental questions: how should an agent's belief be represented, and how should interactions update these beliefs? Existing multi-agent systems primarily rely on natural-language communication, where agents exchange, critique, and revise explicit reasoning traces through dialogue~\citep{du2024debate, liang2024divergent, chen2024reconcile}. Recent advances in latent reasoning have explored continuous representations for intermediate computation, either by recurrently feeding back hidden states~\citep{hao2024coconut, shen2025codi, geiping2025recurrent} or by constructing mixture token embeddings~\citep{zhang2025softthinking, shi2025swireasoning, xu2025softcot}. Extending this direction to multi-agent systems, latent communication methods enable agents to exchange internal representations through activations, hidden states, or key--value caches~\citep{ramesh2025activations, ye2025kvcomm, fu2025c2c}, with recent frameworks supporting latent collaboration through shared working memories~\citep{zou2025latentmas}, learned communication prefixes~\citep{zheng2025thoughtcomm}, or trained receivers~\citep{du2025interlat}. Together, these advances establish continuous representations as a medium for reasoning and communication, motivating a further question: 
\begin{center}
\vspace{0pt}
\textbf{\emph{Can collective intelligence emerge from interacting latent beliefs?}}
\vspace{0pt}
\end{center}

\begin{figure}[h]
\centering
\includegraphics[width=\linewidth]{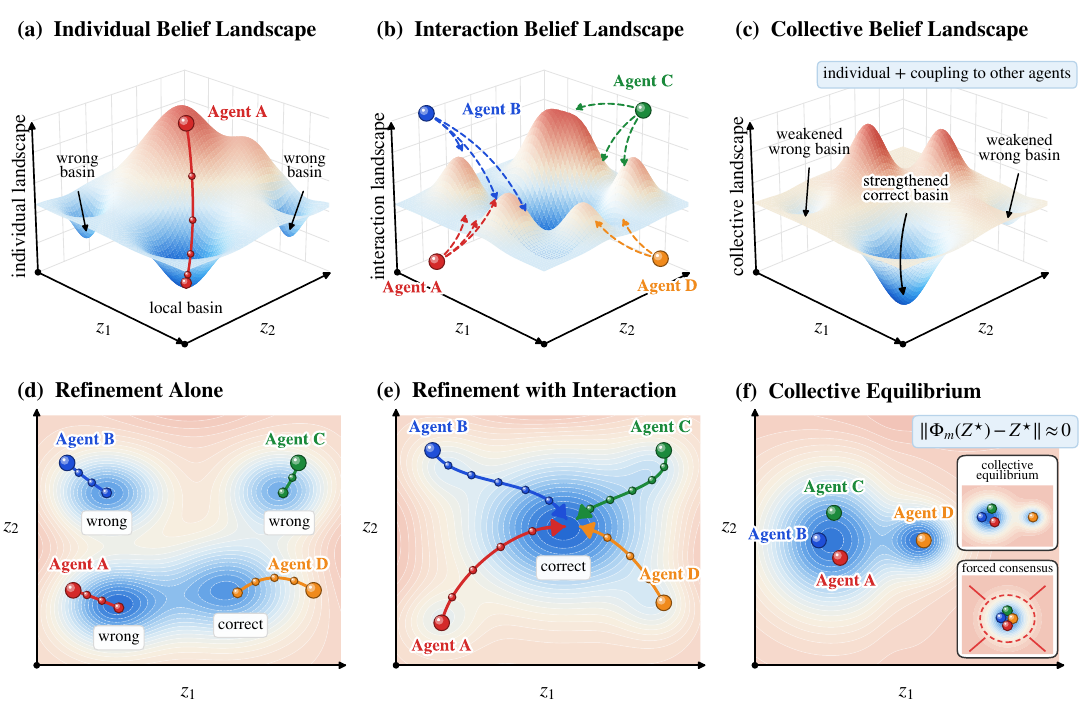}
\caption{\footnotesize\textbf{Illustration of multi-agent communication from a belief dynamics view.} Schematic: the landscapes are illustrative; the axes are two coordinates of the latent belief space, and the basins are basins of attraction of the belief update.}
\label{fig:teaser}
\end{figure}

Motivated by this perspective, we formulate multi-agent reasoning as the evolution of a joint belief state and introduce \textbf{Stream of Belief} (\textbf{SoB}). 
We propose to construct \emph{Belief Tokens}, an agentic communication medium, to facilitate the belief propagation of multi-agent systems in high-dimensional latent space. Additionally, in order to push forward the joint belief state, we construct two complementary processes: \emph{local refinement}, through which each agent independently refines its beliefs, and \emph{global interaction}, through which information propagates across agents. The interplay between local and global operations allows individual belief trajectories to continually reshape one another.

Stream of Belief iterates the joint belief state toward a stable equilibrium, at which further interaction induces no meaningful information propagation, and decodes a collective solution from the resulting latent state. Equilibrium does not imply consensus: agents may retain heterogeneous beliefs while achieving global stability, as in classical influence models whose stable states retain persistent disagreement \citep{abelson1964mathematical, friedkin1990social, acemoglu2013opinion, bindel2015how}.

\underline{Our contributions are threefold.} 
\textbf{(1)}, we formulate collective reasoning as the evolution of interacting latent beliefs toward equilibrium. 
\textbf{(2)}, we propose Stream of Belief, a multi-agent collaboration framework that outperforms the strongest latent reasoning baselines, (i) improving accuracy by up to 15.7 points, (ii) reducing text-token usage by 51\% on average (up to 82\%), and (iii) delivering 1.1$\times$--2.6$\times$ lower token latency than the most efficient latent baselines.
\textbf{(3)}, through extensive experiments, we demonstrate the generalization of our method across tasks and models.

\clearpage
\newpage

\section{Preliminaries and Notations}
\label{sec:prelim}

\textbf{Language model.}
Let $f_\theta$ denote a frozen autoregressive language model
with vocabulary $\mathcal V$ and input embedding matrix $E$,
where $E[v]\in\mathbb R^d$ is the embedding of token $v$.
Given an input sequence, the model produces a last-layer
hidden state $h$ and predicts the next-token distribution
$p=\operatorname{softmax}(W_{\mathrm{out}}h)$.
The autoregressive generation selects a discrete
token from $p$ and feeds its embedding back as the next input.

\textbf{Belief tokens.}
\label{sec:belief_tokens}
Following the probability-weighted embedding formulation
of Soft Thinking~\citep{zhang2025softthinking},
we represent each generation step by a continuous mixture
of token embeddings. Given the next-token distribution
$p_\ell$ at generation step $\ell$, the \emph{belief token} is
\begin{equation}
z_\ell = \sum_{v\in\mathcal V} p_\ell(v)E[v].
\label{eq:belief_token}
\end{equation}
Each belief token is fed back as the next input embedding.\footnote{In practice, we restrict $p_\ell$ to its top-50 tokens and renormalize the probabilities before computing the weighted sum.}
Repeating this process for $L$ autoregressive steps yields
a belief block $z=(z_1,\ldots,z_L)\in\mathbb R^{L\times d}$.
By construction, belief tokens lie in the convex hull of
vocabulary embeddings and can be fed directly into the model
without a learned projection. We examine their distributional
properties in Section~\ref{sec:exp_distribution}.

\begin{tcolorbox}[
    title={Example: Expressiveness of Belief Tokens},
    colback=black!2,
    colframe=softblue,
    colbacktitle=softblue,
    coltitle=white,
    fonttitle=\bfseries,
    boxrule=1.5pt,
    arc=2pt,
    left=8pt, right=8pt, top=6pt, bottom=6pt
]
Consider a multiple-choice question where the next-token
distribution assigns probabilities to four answer labels
$A$, $B$, $C$, and $D$. Suppose the normalized weights are
\[
\bigl(p(A),p(B),p(C),p(D)\bigr)
=(0.55,\,0.30,\,0.15,\,0.00).
\]
Conventional greedy decoding selects $A$ and feeds back
its embedding $E[A]$. In contrast, a belief token represents
a probability-weighted combination of multiple candidates:
\[
z=0.55\,E[A]+0.30\,E[B]+0.15\,E[C]+0.00\,E[D].
\]
Our continuous
representation incorporates contributions from multiple
candidate embeddings. As agents exchange and refine their
beliefs, subsequent token distributions may change,
allowing different candidates to influence the evolving
latent representations.
\end{tcolorbox}

\textbf{Agents and beliefs.}
We consider $N$ agents $A_1,\ldots,A_N$ instantiated from
the same frozen model $f_\theta$ with agent-specific personas.\footnote{Each persona is a one-sentence system prompt, e.g.\ ``You are skeptical and try to rule out wrong options.''(See \S~\ref{app:prompts}).}
We write $A_i(c)\in\mathbb R^{L\times d}$ for the block of
$L$ belief tokens generated autoregressively by agent $i$
from context $c$ using Eq.~\ref{eq:belief_token}.
The context combines text inputs (the problem and
instructions) with continuous belief blocks inserted
directly as input embeddings
(Figure~\ref{fig:alg_demo}).
We refer to each agent's current latent reasoning state
$z_i\in\mathbb R^{L\times d}$ as its \emph{belief}.
The collection of agent beliefs forms the joint state
$Z=(z_1,\ldots,z_N)$.
Reasoning proceeds in cycles indexed by $k$, with
$z_i^{(k)}$ and $Z^{(k)}$ denoting the states at the start
of cycle $k$; the cycle index is omitted when describing
an individual update.

\section{Stream of Belief}
\label{sec:lat_rep}

Each agent initializes its belief from the problem $x$ as
$z_i \leftarrow A_i(x)$. The beliefs are subsequently
updated through two complementary operations:

\textbf{Global Interaction.}
Each agent updates its belief by incorporating the current
beliefs of other agents alongside its own, enabling
information to propagate across the system:
\begin{equation}
[\mathsf G(Z)]_i =
A_i\bigl(x, z_i, \{z_j\}_{j\neq i}\bigr).
\label{eq:global}
\end{equation}

\begin{figure}[!h]
\centering
\includegraphics[width=\linewidth]{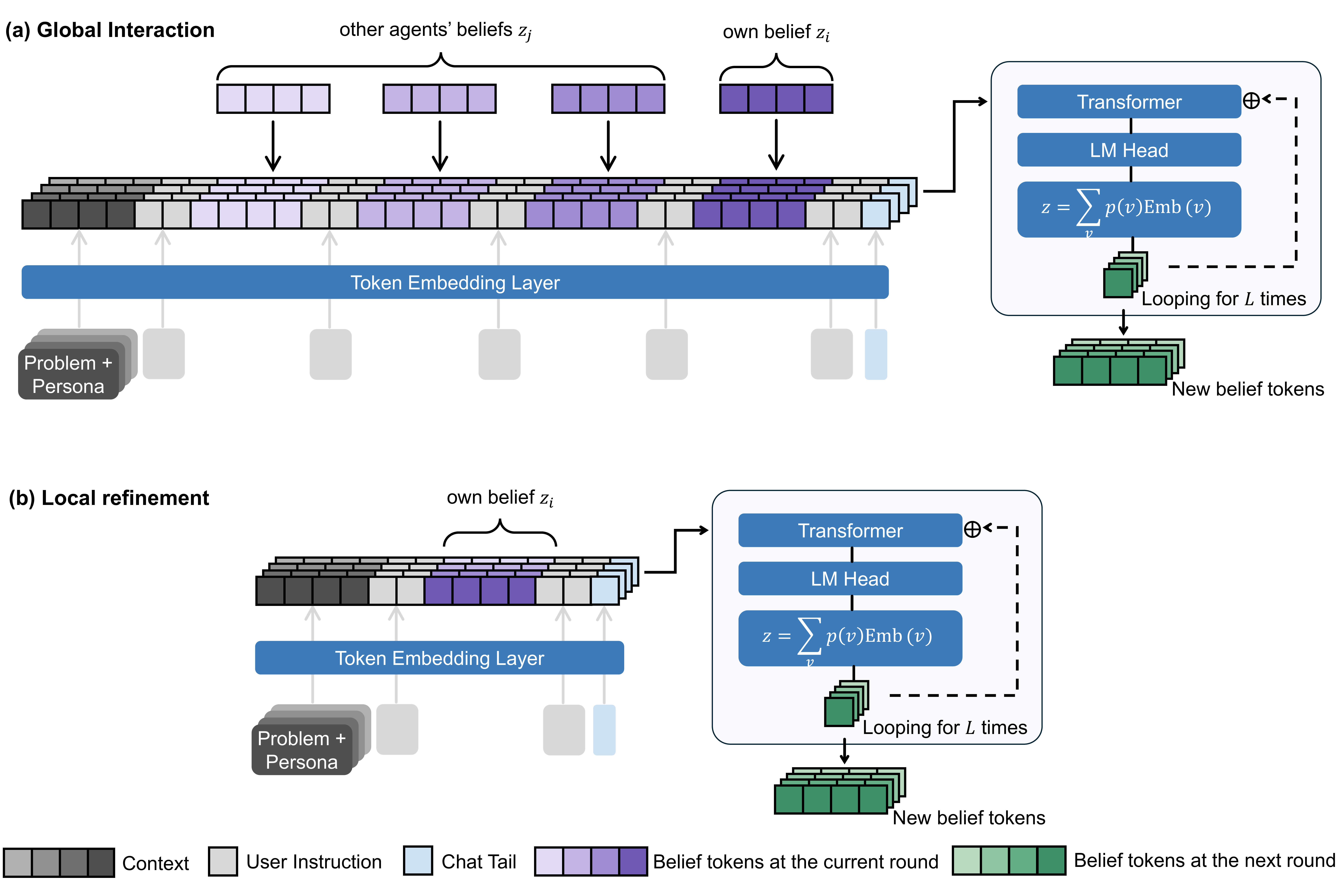}
\caption{\footnotesize
\textbf{Stream of Belief at the embedding layer.}
(a) Global interaction conditions each agent on its own and other agents' beliefs, while (b) local refinement uses only its own belief.
Both operations autoregressively generate $L$ continuous belief tokens via probability-weighted token embeddings (Eq.~\ref{eq:belief_token}), replacing the previous belief block.
}
\label{fig:alg_demo}
\end{figure}

\textbf{Local Refinement.}
Each agent independently refines its belief by conditioning
only on its previous belief and the problem:
\begin{equation}
[\mathsf L(Z)]_i =
A_i\bigl(x, z_i\bigr).
\label{eq:local}
\end{equation}
Both operations employ the same belief-generation procedure,
differing only in the information available to each agent.
All agents update synchronously, with newly generated belief
blocks replacing their previous states while maintaining a
fixed size of $L\times d$ (Figure~\ref{fig:alg_demo}).
Each update reconstructs its context from the problem and
current beliefs, preventing the accumulation of intermediate
states across cycles.

\begin{wrapfigure}{r}{0.5\textwidth}
\vspace{1pt}
\footnotesize
\captionsetup[algorithm]{font=footnotesize}
\begin{minipage}[t]{\linewidth}\raggedright
\algrenewcommand\algorithmiccomment[1]{\hfill{\color{gray}\(\triangleright\)~#1}}
\hrule height 1.5pt
\vspace{3pt}
\captionof{algorithm}{\textbf{Stream of Belief}}
\label{alg:beliefstream}
\vspace{-7pt}
\hrule height 0.4pt
\vspace{4pt}
\begin{algorithmic}[1]
\Require problem $x$; agents $\{A_i\}_{i=1}^N$;
depth $m$; patience $P$; maximum cycles $K_{\max}$
\State $Z\gets(A_1(x),\ldots,A_N(x))$
\For{$k=1,\ldots,K_{\max}$}
    \State $Z\gets\Phi_m(Z)$ \Comment{belief iteration}
    \State $\hat y^{(k)}
    \gets\operatorname{Maj}_i
    \operatorname{Decode}(x,z_i)$ \Comment{short probes}
    \If{$k\ge P$ \textbf{and}
    $\hat y^{(k-P+1)}=\cdots=\hat y^{(k)}$}
        \State \textbf{break} \Comment{early stopping}
    \EndIf
\EndFor
\State \Return $\operatorname{Maj}_i
\operatorname{Decode}(x,z_i)$ \Comment{full solutions}
\end{algorithmic}
\vspace{1pt}
\hrule height 1.5pt
\end{minipage}
\vspace{-10pt}
\end{wrapfigure}

\textbf{Belief Update.}
We organize collective reasoning into \emph{cycles}, each
consisting of one global interaction followed by $m$ local
refinements:
\begin{equation}
Z^{(k+1)} = \Phi_m\bigl(Z^{(k)}\bigr),
\qquad
\Phi_m = \mathsf L^m \circ \mathsf G.
\label{eq:cycle}
\end{equation}
Here, $m$ controls the depth of independent refinement
between successive interactions, with $m=0$ corresponding
to interaction-only updates. We refer to the repeated
application of $\Phi_m$ as \emph{belief iteration}.

\textbf{Collective Equilibrium and Inference.}
Through repeated belief iteration, the joint state evolves
under $\Phi_m$. A collective equilibrium is a fixed point
$Z^\star=\Phi_m(Z^\star)$, where further interaction leaves
the joint state unchanged. We characterize convergence
through the relative belief residual between successive
cycles:
\begin{equation}
r_k =
\frac{\|Z^{(k+1)}-Z^{(k)}\|_F}
{\|Z^{(k)}\|_F+\delta},
\end{equation}
where $\delta>0$ ensures numerical stability.

%During inference, agents exchange and refine beliefs without decoding intermediate textual messages.

\textbf{Intermediate Decoding and Early Stopping.}
We denote by $\operatorname{Decode}(x,z_i)$ the textual
answer decoded by agent $i$ conditioned on problem $x$
and its belief $z_i$, and by $\operatorname{Maj}_i$ the
majority vote over agents' decoded answers.
After each cycle $k$, agents decode short probe answers,
whose majority defines the collective probe $\hat y^{(k)}$.
These probes are used only for monitoring and do not
re-enter belief updates. Inference stops when the collective
probe remains unchanged for $P$ consecutive cycles
(patience) or reaches $K_{\max}$ cycles (maximum budget).
Finally, each agent decodes a full solution from its
final belief, and the collective prediction is obtained
by majority voting. Probe and final decoding use the
same model with different prompts and token budgets
(Appendix~\ref{app:prompts}); Algorithm~\ref{alg:beliefstream} summarizes the full procedure.

%Figures~\ref{fig:discretization} and~\ref{fig:trajectories} use this residual to analyze the belief dynamics. Importantly, equilibrium requires stability of the collective state, not consensus among agents, which may retain distinct beliefs (Figure~\ref{fig:discretization}b).

\begin{figure}[!h]
\centering
\includegraphics[width=\linewidth]{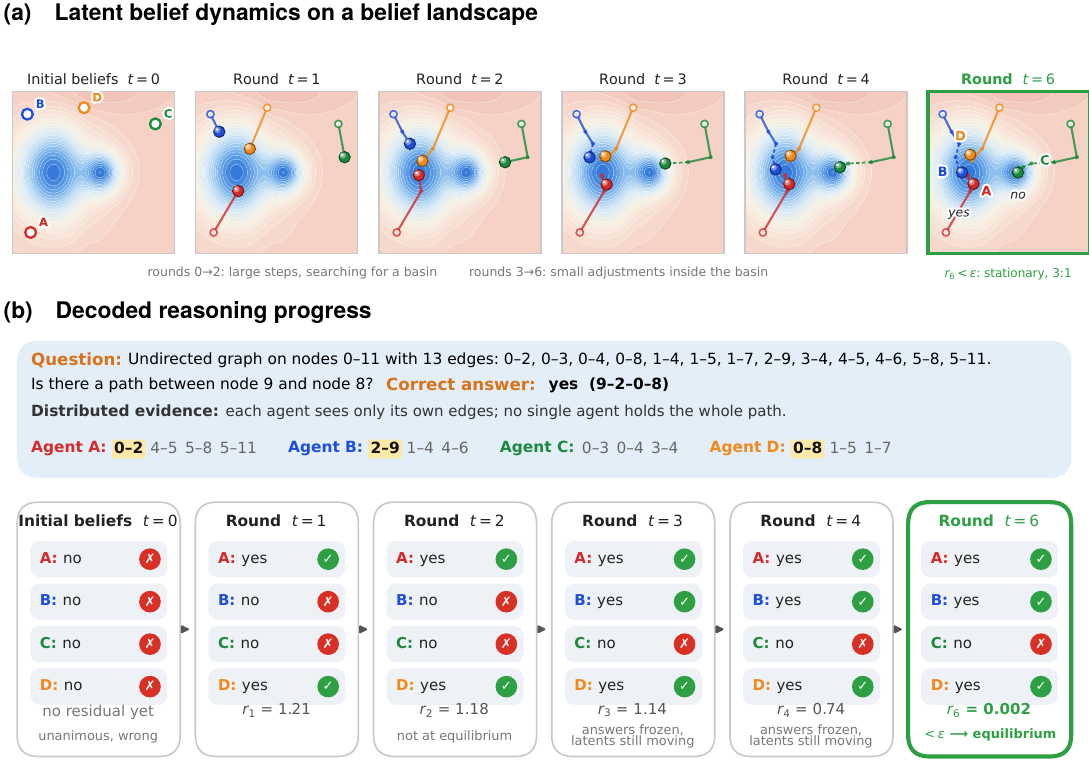}
\caption{\footnotesize \textbf{Qualitative demonstration of reaching a collective equilibrium.} (a) Schematic: the beliefs of four agents on a two-dimensional belief landscape, drawn so that each belief moves in the cycles in which its decoded answer in (b) changes (hollow markers: initial beliefs). In cycles $0$--$2$ the agents take large steps towards an attractor, and in cycles $3$--$6$ they make small adjustments inside its basin. Agents A, B, and D settle in the basin decoded as ``yes'' and agent C in the neighboring basin decoded as ``no''. (b) A graph-connectivity problem with distributed evidence (\texttt{Qwen3-8B}, $N=4$; each agent observes only its own edges). The decoded answers are shown at every cycle together with the belief residual $r_k$. The decoded answers stop changing after cycle~3, while the latent beliefs keep moving until cycle~6, where $r_6<\epsilon$: the joint state is stationary although the agents do not agree (3:1), and the collective answer is decoded from this equilibrium.}
\label{fig:discretization}
\end{figure}

\section{Experiments}
\label{sec:experiments}

Our experiments investigate four questions:
\textbf{1.} How well does SoB generalize across reasoning tasks, model families, and model scales (\S~\ref{sec:exp_main})?
\textbf{2.} How does SoB trade off reasoning accuracy against inference latency and token usage (\S~\ref{sec:exp_efficiency})?
\textbf{3.} How do latent belief representations evolve relative to the model's native embedding space (\S~\ref{sec:exp_distribution})?
\textbf{4.} How do belief capacity, agent population, and iteration dynamics affect collective reasoning (\S~\ref{sec:exp_ablation})?

\subsection{Experimental Setup}
\label{sec:exp_setup}

\textbf{Base models.}
We use one instruction-tuned model from each of four families:
\texttt{Qwen3-8B} (non-thinking mode)~\citep{yang2025qwen3}, \texttt{Llama-3.1-8B-Instruct}~\citep{grattafiori2024llama3}, \texttt{Gemma-3-4B-it}~\citep{gemma2025gemma3}, and \texttt{Phi-4-mini-instruct}~\citep{abouelenin2025phi4mini}.
For the scaling study (Table~\ref{tab:scaling}) we additionally use the Qwen3 dense series from 0.6B to 32B and the Llama family, \texttt{Llama-3.2-1B/3B-Instruct}, \texttt{Llama-3.1-8B-Instruct}, \texttt{Llama-3.3-70B-Instruct} and \texttt{Llama-3.1-405B-Instruct}.

\textbf{Benchmarks.}
Eight benchmarks cover mathematical, scientific, commonsense, logical, and multi-hop reasoning:
GSM8K~\citep{cobbe2021gsm8k}, MATH-500~\citep{hendrycks2021math,lightman2024verify}, MMLU-Pro~\citep{wang2024mmlupro}, GPQA-Diamond~\citep{rein2024gpqa}, ARC-Challenge~\citep{clark2018arc}, MedQA~\citep{jin2021medqa}, BIG-Bench Hard~\citep{suzgun2023bbh}, and HotpotQA~\citep{yang2018hotpotqa}.

\textbf{Protocol.}
$N{=}4$ agents share the frozen base model and update synchronously.
Each belief $z_i$ is a block of $L{=}384$ belief tokens.
Each cycle is one global interaction followed by $m{=}1$ local refinement; we run at most $K_{\max}{=}16$ cycles and stop once the collective probe is unchanged for $P{=}2$ cycles.
Each agent then decodes a solution of up to 512 tokens from its own belief, and $\hat y$ is the majority over parsable answers.
%The belief residual $r_k$, step displacements $\Delta_G$ and $\Delta_L^{(s)}$, and inter-agent disagreement are logged for analysis only.

\textbf{Baselines.}
We compare with six latent reasoning methods.
LatentMAS~\citep{zou2025latentmas}, which passes latent thoughts through a shared working memory without training;
ThoughtComm~\citep{zheng2025thoughtcomm}, which extracts latent thoughts from the agents' hidden states with a sparse autoencoder and injects them into the other agents as trained prefixes; Interlat~\citep{du2025interlat}, which trains a receiver to read the last hidden states of a sender;
SwiReasoning~\citep{shi2025swireasoning} and Soft Thinking~\citep{zhang2025softthinking}, both training-free,
and SoftCoT~\citep{xu2025softcot}, which trains a projection from the soft thoughts of a small assistant model into the embedding space of the frozen base model.
We also compare with \emph{multi-agent debate}~\citep{du2024debate}.
Implementation details of all baselines are given in Appendix~\ref{app:baselines}.

\subsection{Accuracy}
\label{sec:exp_main}

\begin{table}[H]
\centering
\caption{\footnotesize Accuracy (\%) against baselines. Bold and underlined values indicate the best and second-best among latent methods.}
\label{tab:main}
\renewcommand{\arraystretch}{1.11}
\resizebox{\textwidth}{!}{%
\begin{tabular}{cl|cccc|cccc}
\toprule[1.5pt]
\raisebox{-1.5ex}{\textbf{Model}} & \raisebox{-1.5ex}{\textbf{Method}} & \multicolumn{4}{c|}{\textbf{Math and science}} & \multicolumn{4}{c}{\textbf{Knowledge, logic, and multi-hop}}\\
\cmidrule(lr){3-6}\cmidrule(lr){7-10}
&& \textbf{GSM8K} & \textbf{MATH-500} & \textbf{MMLU-Pro} & \textbf{GPQA-D} & \textbf{ARC-C} & \textbf{MedQA} & \textbf{BBH} & \textbf{HotpotQA}\\
\midrule
\multirow{8}{*}{\rotatebox{90}{Qwen3-8B}}
& Multi-agent debate & 96.5\std{1.3} & 86.4\std{2.2} & 75.3\std{2.5} & 53.0\std{3.4} & 94.5\std{1.6} & 73.7\std{2.6} & 93.5\std{1.2} & 70.7\std{3.6}\\
\midrule
& LatentMAS & \underline{87.5\std{2.3}} & \underline{65.6\std{3.1}} & \underline{62.7\std{2.7}} & \underline{34.3\std{3.4}} & \underline{91.0\std{2.1}} & \underline{64.0\std{2.8}} & \underline{86.5\std{1.6}} & \underline{60.0\std{3.9}}\\
& ThoughtComm & 43.5\std{3.5} & 5.2\std{1.4} & 16.0\std{2.2} & 13.1\std{2.5} & 69.5\std{3.3} & 33.3\std{2.7} & 24.8\std{2.0} & 48.7\std{4.2}\\
& Interlat & 84.5\std{2.5} & 47.2\std{3.2} & 25.7\std{2.4} & 13.1\std{2.4} & 85.0\std{2.5} & 52.7\std{3.0} & 56.8\std{2.4} & 59.3\std{4.0}\\
& SwiReasoning & 23.0\std{2.9} & 25.6\std{2.6} & 21.0\std{2.3} & 7.6\std{1.9} & 19.5\std{2.7} & 18.3\std{2.2} & 25.8\std{2.2} & 16.7\std{3.1}\\
& Soft Thinking & 5.5\std{1.6} & 23.2\std{2.7} & 7.7\std{1.5} & 1.5\std{0.9} & 11.0\std{2.2} & 4.3\std{1.2} & 5.5\std{1.1} & 30.0\std{3.7}\\
& SoftCoT & 39.0\std{3.5} & 26.0\std{2.7} & 13.0\std{2.0} & 9.6\std{2.1} & 14.5\std{2.5} & 8.3\std{1.6} & 24.8\std{1.8} & 17.3\std{3.1}\\
& \textbf{Ours} & \textbf{94.5\std{1.6}} & \textbf{77.2\std{2.7}} & \textbf{69.7\std{2.7}} & \textbf{44.9\std{3.6}} & \textbf{92.5\std{1.8}} & \textbf{76.3\std{2.4}} & \textbf{93.5\std{1.2}} & \textbf{61.3\std{3.8}}\\
\midrule[1pt]
\midrule[1pt]
\multirow{8}{*}{\rotatebox{90}{Phi-4-mini}}
& Multi-agent debate & 88.5\std{2.2} & 58.4\std{3.2} & 55.7\std{2.9} & 32.8\std{3.2} & 87.0\std{2.5} & 55.0\std{3.0} & 81.2\std{2.0} & 24.0\std{3.5}\\
\midrule
& LatentMAS & 73.5\std{3.0} & 11.2\std{1.9} & \underline{38.3\std{2.7}} & \underline{23.7\std{3.0}} & \underline{79.0\std{3.0}} & 25.3\std{2.5} & 62.0\std{2.4} & \underline{46.0\std{4.0}}\\
& ThoughtComm & 61.0\std{3.5} & 37.2\std{3.0} & 25.3\std{2.5} & 16.7\std{2.7} & 36.5\std{3.2} & 34.0\std{2.8} & 42.8\std{2.4} & 21.3\std{3.3}\\
& Interlat & 68.5\std{3.3} & 32.0\std{2.9} & 24.7\std{2.5} & 21.2\std{2.8} & 77.5\std{2.9} & 38.7\std{2.8} & 56.2\std{2.4} & \textbf{50.0\std{4.2}}\\
& SwiReasoning & \underline{83.5\std{2.6}} & \underline{51.6\std{3.2}} & 37.7\std{2.7} & 11.6\std{2.3} & 70.0\std{3.1} & \underline{54.0\std{2.9}} & 63.5\std{2.3} & 20.0\std{3.3}\\
& Soft Thinking & 77.0\std{2.9} & 36.0\std{3.0} & 30.7\std{2.8} & 7.6\std{1.9} & 71.0\std{3.1} & 38.3\std{3.0} & \underline{71.5\std{2.0}} & 8.7\std{2.3}\\
& SoftCoT & 50.0\std{3.6} & 17.6\std{2.4} & 10.3\std{1.8} & 16.2\std{2.6} & 50.0\std{3.6} & 27.0\std{2.6} & 32.8\std{2.3} & 18.0\std{2.9}\\
& \textbf{Ours} & \textbf{91.0\std{2.1}} & \textbf{56.0\std{3.3}} & \textbf{54.0\std{2.9}} & \textbf{28.3\std{3.2}} & \textbf{90.5\std{2.2}} & \textbf{55.3\std{2.8}} & \textbf{80.2\std{1.8}} & 37.3\std{3.9}\\
\midrule[1pt]
\midrule[1pt]
\multirow{8}{*}{\rotatebox{90}{Llama-3.1-8B}}
& Multi-agent debate & 89.5\std{2.1} & 54.0\std{3.2} & 47.7\std{2.8} & 30.3\std{3.3} & 83.5\std{2.6} & 70.0\std{2.6} & 60.5\std{2.5} & 51.3\std{4.0}\\
\midrule
& LatentMAS & \underline{85.5\std{2.5}} & 46.0\std{3.0} & 40.0\std{3.0} & 22.2\std{3.0} & 79.5\std{2.9} & 68.3\std{2.7} & \underline{66.2\std{2.3}} & \textbf{67.3\std{3.7}}\\
& ThoughtComm & 72.5\std{3.3} & 29.6\std{2.9} & 24.3\std{2.5} & 12.6\std{2.3} & 73.5\std{2.9} & 52.7\std{3.0} & 52.5\std{2.4} & 44.7\std{4.1}\\
& Interlat & 53.5\std{3.6} & 25.6\std{2.7} & 21.3\std{2.3} & 23.2\std{3.0} & 35.0\std{3.5} & 34.0\std{2.7} & 35.8\std{2.4} & 9.3\std{2.4}\\
& SwiReasoning & \textbf{87.0\std{2.3}} & \textbf{49.6\std{3.1}} & \underline{42.7\std{2.8}} & \underline{24.2\std{3.1}} & \underline{82.5\std{2.7}} & \underline{69.0\std{2.7}} & 60.0\std{2.4} & \underline{55.3\std{4.2}}\\
& Soft Thinking & 81.5\std{2.8} & 40.8\std{3.2} & 31.0\std{2.7} & 13.6\std{2.4} & 68.5\std{3.2} & 36.7\std{2.7} & 60.8\std{2.3} & 26.0\std{3.5}\\
& SoftCoT & 47.0\std{3.4} & 13.6\std{2.1} & 20.7\std{2.3} & 23.7\std{3.0} & 66.0\std{3.3} & 47.3\std{2.8} & 33.2\std{2.3} & 37.3\std{4.0}\\
& \textbf{Ours} & 84.0\std{2.6} & \underline{46.8\std{3.1}} & \textbf{47.7\std{2.9}} & \textbf{31.8\std{3.3}} & \textbf{84.5\std{2.6}} & \textbf{69.3\std{2.7}} & \textbf{70.0\std{2.3}} & 51.3\std{4.1}\\
\midrule[1pt]
\midrule[1pt]
\multirow{8}{*}{\rotatebox{90}{Gemma-3-4B}}
& Multi-agent debate & 89.5\std{2.1} & 80.8\std{2.5} & 47.7\std{2.8} & 32.3\std{3.3} & 78.0\std{2.9} & 45.0\std{2.9} & 89.8\std{1.5} & 54.0\std{4.1}\\
\midrule
& LatentMAS & 83.0\std{2.7} & 58.4\std{3.2} & 31.3\std{2.6} & 9.6\std{2.1} & 82.0\std{2.7} & 25.3\std{2.6} & \textbf{84.8\std{1.6}} & 33.3\std{3.8}\\
& ThoughtComm & 78.5\std{3.0} & 39.6\std{3.2} & 28.3\std{2.6} & 16.7\std{2.6} & 65.0\std{3.3} & 44.0\std{2.9} & 70.0\std{2.1} & 42.0\std{4.0}\\
& Interlat & 60.0\std{3.5} & 28.0\std{2.7} & 12.7\std{1.9} & 20.7\std{2.8} & 72.5\std{3.2} & 31.3\std{2.7} & 64.0\std{2.4} & 38.0\std{4.0}\\
& SwiReasoning & 81.5\std{2.6} & \textbf{73.2\std{2.7}} & 30.3\std{2.7} & 15.7\std{2.6} & 81.5\std{2.8} & \underline{51.3\std{2.8}} & 82.0\std{1.8} & \textbf{63.3\std{4.1}}\\
& Soft Thinking & \underline{84.5\std{2.5}} & \underline{67.6\std{3.0}} & 33.7\std{2.8} & 15.7\std{2.6} & \underline{83.0\std{2.6}} & 49.0\std{2.8} & 81.0\std{1.9} & \underline{60.0\std{4.0}}\\
& SoftCoT & 53.0\std{3.5} & 44.4\std{3.1} & \underline{36.3\std{2.8}} & \underline{23.7\std{3.0}} & 57.5\std{3.5} & 46.7\std{2.9} & 54.2\std{2.4} & 26.7\std{3.6}\\
& \textbf{Ours} & \textbf{89.0\std{2.2}} & 66.0\std{3.0} & \textbf{47.0\std{3.0}} & \textbf{31.3\std{3.3}} & \textbf{84.5\std{2.5}} & \textbf{54.7\std{2.8}} & \underline{83.0\std{1.8}} & 33.3\std{3.8}\\
\bottomrule[1.5pt]
\end{tabular}}
\end{table}

\takeaway{Interacting latent beliefs provide an effective and generalizable mechanism for collective reasoning across tasks and model families.}

Table~\ref{tab:main} shows that collective reasoning can be effectively organized through continuous latent interaction, consistently outperforming latent baselines across diverse tasks and models.
Its competitive performance with text-based debate suggests effective collaboration can also emerge through latent interaction.

\begin{figure}[!h]
\centering
\includegraphics[width= 0.8\linewidth]{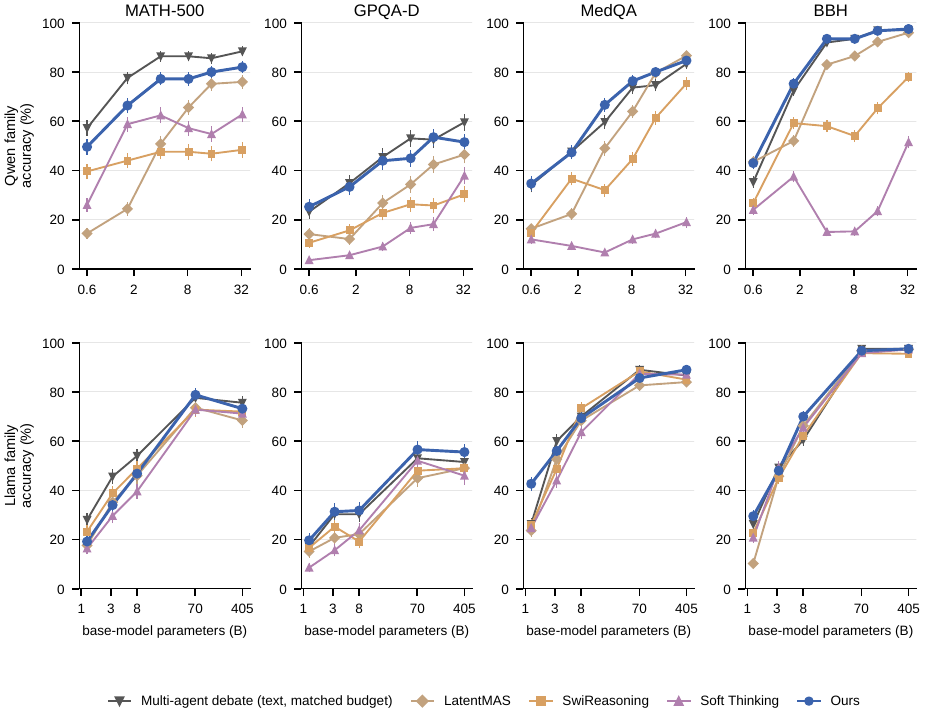}
\caption{\footnotesize \textbf{Scaling with model size.} Accuracy across benchmarks for the Qwen3 family (top) and Llama family (bottom).}
\label{fig:scaling}
\end{figure}

\begin{table}[H]
\centering
\caption{\footnotesize \textbf{Accuracy (\%) across model scales.}}
\label{tab:scaling}
\renewcommand{\arraystretch}{0.93}
\begin{minipage}[t]{0.49\textwidth}\centering
\resizebox{\linewidth}{!}{%
\begin{tabular}[t]{ll|cccc}
\toprule[1.5pt]
\textbf{Model} & \textbf{Method} & \textbf{MATH-500} & \textbf{GPQA-D} & \textbf{MedQA} & \textbf{BBH}\\
\midrule
\multirow{5}{*}{Qwen3-0.6B}
& Multi-agent debate & 57.2\std{3.2} & 23.2\std{2.9} & 34.0\std{2.9} & 35.2\std{2.3}\\
\midrule
& LatentMAS & 14.4\std{2.1} & \underline{14.1\std{2.4}} & \underline{16.3\std{2.1}} & \textbf{43.5\std{2.5}}\\
& SwiReasoning & \underline{39.6\std{3.1}} & 10.6\std{2.2} & 14.7\std{2.1} & 26.7\std{2.2}\\
& Soft Thinking & 26.0\std{2.8} & 3.5\std{1.3} & 12.0\std{2.0} & 24.0\std{2.1}\\
& \textbf{Ours} & \textbf{49.6\std{3.1}} & \textbf{25.3\std{3.2}} & \textbf{34.7\std{2.9}} & \underline{43.0\std{2.5}}\\
\midrule[1pt]
\multirow{5}{*}{Qwen3-1.7B}
& Multi-agent debate & 77.6\std{2.6} & 34.8\std{3.4} & 47.7\std{2.8} & 72.5\std{2.3}\\
\midrule
& LatentMAS & 24.4\std{2.8} & 12.1\std{2.4} & 22.3\std{2.4} & 52.0\std{2.5}\\
& SwiReasoning & 44.0\std{3.1} & \underline{15.7\std{2.6}} & \underline{36.7\std{2.7}} & \underline{59.2\std{2.5}}\\
& Soft Thinking & \underline{58.8\std{3.1}} & 5.6\std{1.6} & 9.3\std{1.7} & 37.5\std{2.3}\\
& \textbf{Ours} & \textbf{66.4\std{3.1}} & \textbf{33.3\std{3.4}} & \textbf{47.3\std{2.9}} & \textbf{75.2\std{2.3}}\\
\midrule[1pt]
\multirow{5}{*}{Qwen3-4B}
& Multi-agent debate & 86.4\std{2.2} & 45.5\std{3.5} & 59.7\std{2.9} & 92.0\std{1.3}\\
\midrule
& LatentMAS & 50.8\std{3.1} & \underline{26.8\std{3.1}} & \underline{49.0\std{3.0}} & \underline{83.0\std{1.9}}\\
& SwiReasoning & 47.6\std{3.1} & 22.7\std{3.0} & 32.0\std{2.7} & 58.0\std{2.4}\\
& Soft Thinking & \underline{62.4\std{3.2}} & 9.1\std{2.0} & 6.7\std{1.4} & 15.0\std{1.8}\\
& \textbf{Ours} & \textbf{77.2\std{2.6}} & \textbf{43.9\std{3.7}} & \textbf{66.7\std{2.8}} & \textbf{93.5\std{1.2}}\\
\midrule[1pt]
\multirow{5}{*}{Qwen3-8B}
& Multi-agent debate & 86.4\std{2.1} & 53.0\std{3.4} & 73.7\std{2.6} & 93.5\std{1.3}\\
\midrule
& LatentMAS & \underline{65.6\std{2.9}} & \underline{34.3\std{3.3}} & \underline{64.0\std{2.8}} & \underline{86.5\std{1.7}}\\
& SwiReasoning & 47.6\std{3.2} & 26.3\std{3.1} & 44.7\std{2.8} & 54.0\std{2.5}\\
& Soft Thinking & 57.2\std{3.0} & 16.7\std{2.6} & 12.0\std{1.9} & 15.3\std{1.8}\\
& \textbf{Ours} & \textbf{77.2\std{2.7}} & \textbf{44.9\std{3.5}} & \textbf{76.3\std{2.4}} & \textbf{93.5\std{1.2}}\\
\midrule[1pt]
\multirow{5}{*}{Qwen3-14B}
& Multi-agent debate & 85.6\std{2.3} & 52.5\std{3.5} & 74.7\std{2.5} & 97.0\std{0.9}\\
\midrule
& LatentMAS & \underline{75.2\std{2.7}} & \underline{42.4\std{3.5}} & \underline{79.7\std{2.3}} & \underline{92.2\std{1.3}}\\
& SwiReasoning & 46.8\std{3.1} & 25.8\std{2.9} & 61.3\std{2.8} & 65.2\std{2.5}\\
& Soft Thinking & 54.8\std{3.2} & 18.2\std{2.7} & 14.3\std{1.9} & 23.5\std{2.2}\\
& \textbf{Ours} & \textbf{80.0\std{2.6}} & \textbf{53.5\std{3.5}} & \textbf{80.0\std{2.2}} & \textbf{96.8\std{0.9}}\\
\midrule[1pt]
\multirow{5}{*}{Qwen3-32B}
& Multi-agent debate & 88.4\std{2.0} & 59.6\std{3.5} & 83.3\std{2.2} & 97.0\std{0.8}\\
\midrule
& LatentMAS & \underline{76.0\std{2.7}} & \underline{46.5\std{3.6}} & \textbf{86.7\std{1.9}} & \underline{96.0\std{1.0}}\\
& SwiReasoning & 48.4\std{3.2} & 30.3\std{3.3} & 75.3\std{2.5} & 78.0\std{2.1}\\
& Soft Thinking & 62.8\std{3.1} & 37.9\std{3.5} & 19.0\std{2.2} & 51.5\std{2.5}\\
& \textbf{Ours} & \textbf{82.0\std{2.4}} & \textbf{51.5\std{3.5}} & \underline{84.7\std{2.1}} & \textbf{97.5\std{0.8}}\\
\bottomrule[1.5pt]
\end{tabular}}
\end{minipage}\hfill
\begin{minipage}[t]{0.49\textwidth}\centering
\resizebox{\linewidth}{!}{%
\begin{tabular}[t]{ll|cccc}
\toprule[1.5pt]
\textbf{Model} & \textbf{Method} & \textbf{MATH-500} & \textbf{GPQA-D} & \textbf{MedQA} & \textbf{BBH}\\
\midrule
\multirow{5}{*}{Llama-3.2-1B}
& Multi-agent debate & 28.0\std{2.8} & 17.2\std{2.7} & 26.3\std{2.5} & 26.2\std{2.2}\\
\midrule
& LatentMAS & 17.6\std{2.4} & 15.2\std{2.6} & 23.7\std{2.4} & 10.2\std{1.5}\\
& SwiReasoning & \textbf{23.2\std{2.6}} & \underline{16.7\std{2.7}} & \underline{26.0\std{2.5}} & \underline{22.8\std{2.1}}\\
& Soft Thinking & 16.4\std{2.3} & 8.6\std{2.0} & 24.7\std{2.4} & 20.8\std{2.0}\\
& \textbf{Ours} & \underline{19.2\std{2.5}} & \textbf{19.7\std{2.8}} & \textbf{42.7\std{2.9}} & \textbf{29.5\std{2.3}}\\
\midrule[1pt]
\multirow{5}{*}{Llama-3.2-3B}
& Multi-agent debate & 45.6\std{3.1} & 30.3\std{3.2} & 60.0\std{2.8} & 49.2\std{2.5}\\
\midrule
& LatentMAS & \underline{35.2\std{3.0}} & 20.7\std{2.8} & \underline{52.7\std{2.9}} & 45.8\std{2.5}\\
& SwiReasoning & \textbf{38.8\std{3.1}} & \underline{25.3\std{3.1}} & 48.7\std{2.9} & 45.0\std{2.4}\\
& Soft Thinking & 29.6\std{2.9} & 15.7\std{2.5} & 44.0\std{2.9} & \textbf{49.3\std{2.5}}\\
& \textbf{Ours} & 34.0\std{3.1} & \textbf{31.3\std{3.4}} & \textbf{56.0\std{2.9}} & \underline{48.0\std{2.5}}\\
\midrule[1pt]
\multirow{5}{*}{Llama-3.1-8B}
& Multi-agent debate & 54.0\std{3.0} & 30.3\std{3.1} & 70.0\std{2.6} & 60.5\std{2.4}\\
\midrule
& LatentMAS & 46.0\std{3.4} & 22.2\std{2.9} & 68.3\std{2.6} & \underline{66.2\std{2.3}}\\
& SwiReasoning & \textbf{48.8\std{3.1}} & 19.2\std{2.8} & \textbf{73.3\std{2.5}} & 62.2\std{2.5}\\
& Soft Thinking & 39.6\std{3.1} & \underline{23.7\std{3.1}} & 63.7\std{2.8} & 65.5\std{2.4}\\
& \textbf{Ours} & \underline{46.8\std{3.1}} & \textbf{31.8\std{3.3}} & \underline{69.3\std{2.7}} & \textbf{70.0\std{2.2}}\\
\midrule[1pt]
\multirow{5}{*}{Llama-3.3-70B}
& Multi-agent debate & 77.6\std{2.6} & 53.0\std{3.6} & 89.0\std{1.9} & 97.5\std{0.8}\\
\midrule
& LatentMAS & \underline{73.6\std{2.8}} & 44.9\std{3.5} & 82.7\std{2.1} & \textbf{96.8\std{0.9}}\\
& SwiReasoning & 72.8\std{2.9} & 48.0\std{3.6} & \textbf{88.3\std{1.9}} & \underline{95.8\std{1.0}}\\
& Soft Thinking & 72.8\std{2.7} & \underline{52.0\std{3.5}} & \underline{87.3\std{1.9}} & \underline{95.8\std{1.0}}\\
& \textbf{Ours} & \textbf{78.8\std{2.6}} & \textbf{56.6\std{3.6}} & 85.7\std{2.0} & \textbf{96.8\std{0.9}}\\
\midrule[1pt]
\multirow{5}{*}{Llama-3.1-405B}
& Multi-agent debate & 75.6\std{2.7} & 51.5\std{3.6} & 86.7\std{1.9} & 97.5\std{0.8}\\
\midrule
& LatentMAS & 68.4\std{3.0} & \underline{49.0\std{3.5}} & 84.0\std{2.1} & \underline{97.2\std{0.8}}\\
& SwiReasoning & \underline{72.0\std{2.9}} & \underline{49.0\std{3.6}} & 85.0\std{2.1} & 95.5\std{1.0}\\
& Soft Thinking & 71.2\std{3.0} & 46.0\std{3.6} & \underline{87.0\std{1.9}} & \underline{97.2\std{0.8}}\\
& \textbf{Ours} & \textbf{73.2\std{2.8}} & \textbf{55.6\std{3.4}} & \textbf{89.0\std{1.7}} & \textbf{97.5\std{0.8}}\\
\bottomrule[1.5pt]
\end{tabular}}
\end{minipage}
\end{table}

\takeaway{Latent belief interaction remains effective across model scales.}

\textbf{Scaling with model size.}
Table~\ref{tab:scaling} and Figure~\ref{fig:scaling} demonstrate that SoB maintains competitive performance across the Qwen3 and Llama families without changing its inference configuration.
While text-based debate retains advantages on certain tasks, the consistent effectiveness of SoB across model scales suggests that latent collaboration can benefit from increasingly capable base models.

\begin{figure}[!h]
\centering
\includegraphics[width=\linewidth]{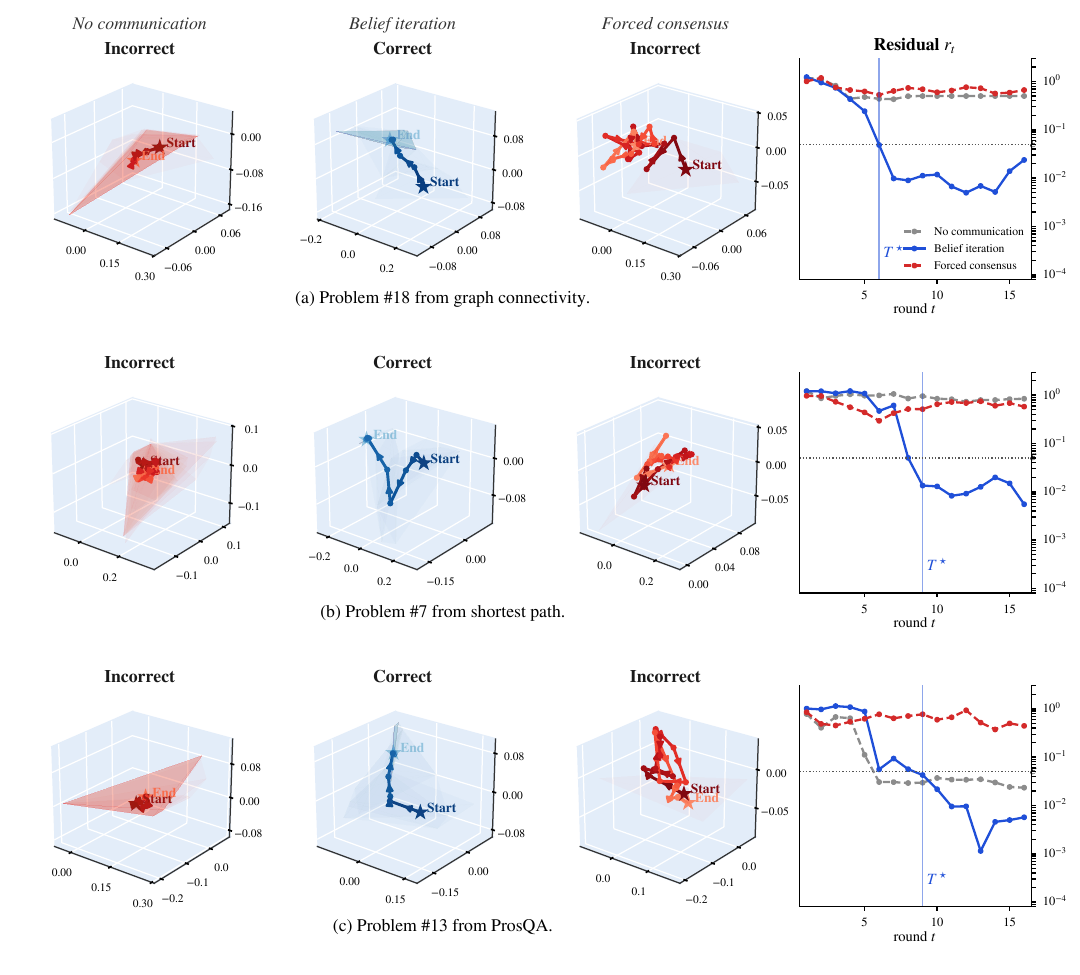}
\caption{\footnotesize \textbf{Qualitative demonstration of latent belief trajectories.} For each problem, the latent beliefs of $N=4$ agents (\texttt{Qwen3-8B}, $L=48$) over cycles $k=0,\ldots,16$ are projected onto 3D with PCA. Lines show the collective trajectory (mean over agents), arrows indicate the progression from one cycle to the next, and the shaded region spans the agents' beliefs at consecutive cycles. Blue and red denote that the collective answer is correct and incorrect, respectively. \emph{No communication}: agents refine their beliefs in isolation; \emph{Belief iteration}: agents interact through their latent beliefs; \emph{Forced consensus}: all beliefs are replaced by their mean after every cycle. The right column shows the belief residual $r_k$; the vertical line marks the equilibrium cycle $\hat k$ at which $r_k<\epsilon$. Examples are from distributed-information tasks in which each agent observes only part of the problem.}
\label{fig:trajectories}
\end{figure}

\subsection{Efficiency}
\label{sec:exp_efficiency}

\begin{figure}[!h]
\centering
\includegraphics[width=\linewidth]{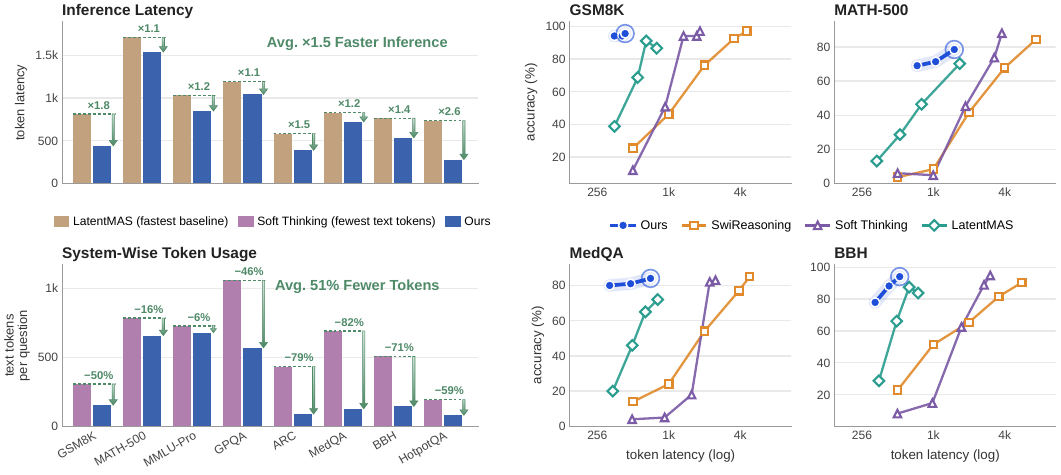}
\caption{\footnotesize \textbf{Latency and token consumption.} (\texttt{Qwen3-8B}). \textbf{Top left:} token latency, the number of tokens generated one after another before the answer. \textbf{Bottom left:} text tokens per question, summed over all agents. \textbf{Right:} accuracy against token latency; our method for belief lengths $L\in\{96,192,384\}$ (ringed: $L=384$), SwiReasoning and Soft Thinking with budgets from 512 to 8{,}192 tokens, and LatentMAS with judger caps of 256, 512, and 1{,}024 tokens.}
\label{fig:efficiency}
\end{figure}

\takeaway{Parallel latent belief interaction enables efficient collective reasoning with reduced sequential latency and explicit text generation.}

\textbf{Inference Latency.}
We measure token latency by sequential generation depth, including both latent and text tokens.
As shown in Figure~\ref{fig:efficiency}, SoB achieves shorter sequential latency than LatentMAS, the fastest baseline, reducing token latency by 1.1$\times$--2.6$\times$ across the eight benchmarks (1.5$\times$ on average) while maintaining competitive reasoning accuracy across different belief lengths.
This efficiency reflects a different organization of collective reasoning: rather than extending sequential reasoning chains, SoB distributes computation across parallel belief trajectories.

\textbf{Text-Token Usage.}
Figure~\ref{fig:efficiency} further shows that SoB substantially reduces explicit text generation compared with Soft Thinking, the baseline with the lowest text-token usage: 51\% fewer text tokens on average, and up to 82\% fewer on individual benchmarks.
By exchanging continuous beliefs instead of textual reasoning traces, SoB decouples inter-agent communication from language generation, reserving text primarily for intermediate probes and final readouts.
These gains reflect reduced sequential depth and textual communication, rather than a reduction in total computational cost.

\subsection{Distributions of Latent Representations}
\label{sec:exp_distribution}

\takeaway{Continuous belief tokens spontaneously align with vocabulary embeddings through collective belief dynamics.}

\begin{figure}[h]
\centering
\includegraphics[width=\linewidth]{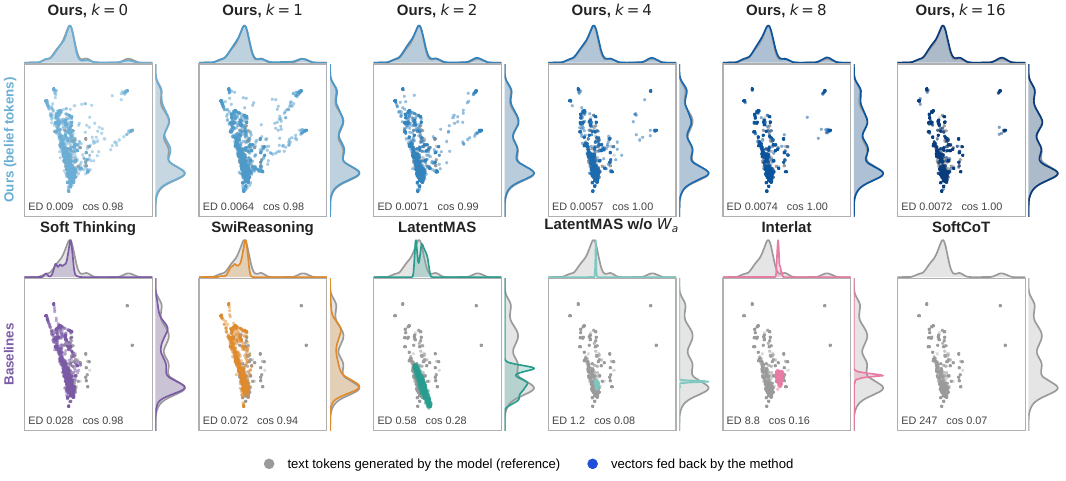}
\caption{\footnotesize \textbf{Latent representation distributions vs text-token distribution.}  \textbf{Top:} Spontaneous alignment of the distribution of our constructed belief tokens. \textbf{Bottom:} the distributions of latent representations from baseline methods. Corner of each panel: energy distance (ED) to the text-token distribution in the full representation space and mean cosine between a vector and its nearest vocabulary embedding (\texttt{Qwen3-8B}, BBH logical deduction).}
\label{fig:dshift}
\end{figure}

\begin{figure}[!h]
\vspace{0pt}
\centering
\begin{minipage}[t]{0.5\linewidth}
    \centering
    \includegraphics[width=\linewidth]{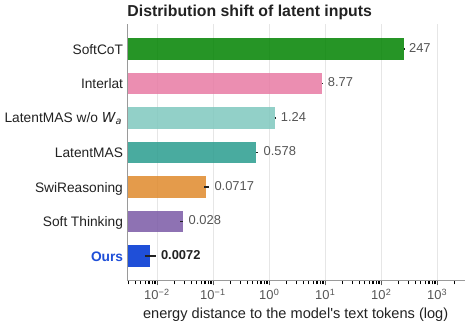}
    \vspace{2pt}
    {\footnotesize \textbf{(a) Distribution shift.}}
\end{minipage}
\hfill
\begin{minipage}[t]{0.49\linewidth}
    \centering
    \includegraphics[width=\linewidth]{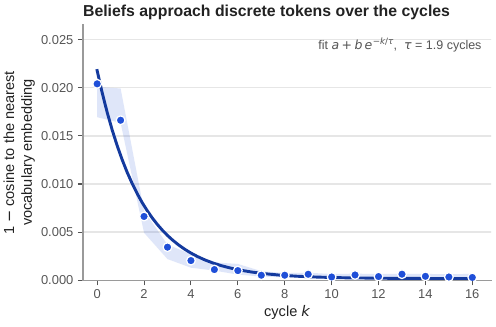}
    \vspace{2pt}
    {\footnotesize \textbf{(b) Spontaneous sharpening.}}
\end{minipage}
\caption{\footnotesize
\textbf{Quantitative demonstration of spontaneous sharpening of belief tokens.}
(a) Energy distance between the latent inputs used by each method and the input embeddings of the tokens generated by the same model in its text chain of thought on the same questions.
(b) Averaged distance ($1-\cos$) between a belief token and its nearest vocabulary embedding across cycles of belief iteration (\texttt{Qwen3-8B}, $N=4$, $L=192$, BBH logical-deduction questions). Belief tokens start as genuine mixtures of multiple token embeddings and rapidly approach individual vocabulary embeddings over successive cycles.
}
\label{fig:belief_geometry}
\end{figure}

\textbf{Spontaneous alignment.}
By construction, belief tokens are probability-weighted mixtures of vocabulary embeddings (Eq.~\ref{eq:belief_token}), allowing multiple candidate tokens to contribute to subsequent computation.
Remarkably, as collective beliefs evolve, these continuous representations spontaneously approach individual vocabulary embeddings (Figure~\ref{fig:belief_geometry}).
Their mean cosine distance to the nearest vocabulary embedding decreases from $0.020$ at $k=0$ to approximately $3\times10^{-4}$ at $k=16$, with a characteristic timescale of $\tau\approx1.9$ cycles.
\emph{This spontaneous alignment suggests that token-level commitment can emerge naturally from continuous belief dynamics, without explicit discretization.}

\textbf{Distributional compatibility.}
To assess how closely latent inputs resemble the model's native token representations, we measure their energy distance (ED)~\citep{szekely2013energy} to embeddings of tokens generated by the same model on the same questions:
\begin{equation}
\mathrm{ED}(X,Y)
=2\,\mathbb E\|X-Y\|
-\mathbb E\|X-X'\|
-\mathbb E\|Y-Y'\|,
\end{equation}
where $X,X'$ and $Y,Y'$ are independent samples from the latent-input and text-token embedding distributions, respectively.
We estimate these expectations using 4{,}000 vectors per distribution in the full embedding space.
Belief tokens remain close to the model's native text-token distribution throughout iteration (Figure~\ref{fig:dshift}), exhibiting substantially smaller distributional shifts than alternative latent reasoning methods (Figure~\ref{fig:belief_geometry}a).

\subsection{Ablations}
\label{sec:exp_ablation}

\begin{figure}[!h]
\centering
\includegraphics[width=\linewidth]{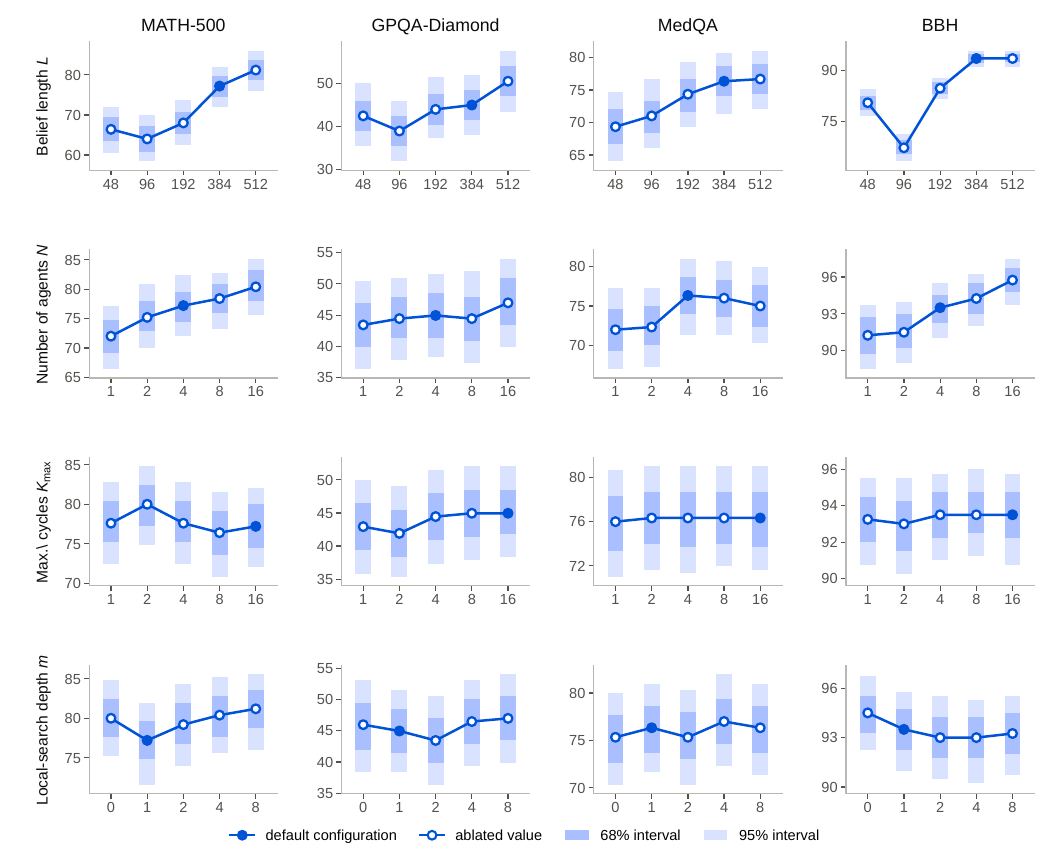}
\caption{\footnotesize \textbf{Ablations of the hyperparameters of SoB} (\texttt{Qwen3-8B}).  Boxes: 68\% (dark) and 95\% (light) bootstrap intervals.}
\label{fig:ablation}
\end{figure}

\takeaway{The capacity to represent beliefs matters more than the extent of collective deliberation.}

\textbf{Quality Factors.}
Figure~\ref{fig:ablation} examines how belief capacity, agent population, interaction cycles, and local-refinement depth shape collective reasoning.
Among these factors, belief capacity exhibits the strongest and most consistent influence, suggesting that richer latent representations provide a more effective foundation for collective reasoning.
In contrast, increasing the number of agents yields task-dependent benefits, indicating that a larger population does not necessarily translate into stronger collective intelligence.

Extending deliberation offers similarly limited benefits.
Increasing the number of interaction cycles or local refinements produces no consistent improvement in accuracy, despite requiring additional belief updates.
Together, these findings suggest that effective collective reasoning depends less on repeatedly revisiting beliefs than on the representational capacity available to each agent.

\section{Related Work}
\label{sec:related}

\textbf{Multi-agent reasoning with language models.}
Recent work extends multi-agent systems to LLM-based agents \citep{guo2024large, tran2025multi, park2023generative, hong2024metagpt, yang2024llm} that collaborate on reasoning, planning, and problem-solving \citep{tao2024magis, wang2025talk, zhao2025sirius}. Existing methods coordinate agents through dialogue and role assignment \citep{yao2022react, wu2024autogen, li2023camel}, structured communication \citep{ye2025kvcomm}, and emergent specialization \citep{mieczkowski2025predicting, huang2025many}. Multi-agent debate improves reasoning and factuality through iterative proposal and critique \citep{du2024debate,liang2024divergent,chan2024chateval}, while general frameworks organize agents into roles and workflows \citep{wu2024autogen,li2023camel,hong2024metagpt,chen2024agentverse,park2023generative}. Other work optimizes interaction graphs \citep{zhuge2024gptswarm}, generates task-specific topologies \citep{zhang2025gdesigner}, and studies scaling through populations \citep{li2024moreagents,qian2025scaling}. 
Our work formulates multi-agent reasoning as the evolution of a joint latent belief state toward a collective equilibrium.

\textbf{Latent interaction between agents.}
Recent MAS have also explored interaction through latent representations \citep{tran2025multi,coda2025exploring,ye2025kvcomm,fu2025c2c,zheng2025thoughtcomm}. Existing work studied intermediate activations \citep{ramesh2025activations}, last-layer hidden states \citep{du2025interlat}, hidden-state deltas \citep{tang2025sde}, or key--value caches \citep{fu2025c2c,shi2025kvcomm, zou2025latentmas, zheng2025thoughtcomm}.
We introduce belief tokens, a belief iteration operator, and a belief residual to characterize the latent states of reasoning agents.

\textbf{Test-time scaling of language model reasoning.}
Chain-of-thought~\citep{wei2022chain}, self-consistency~\citep{wang2023selfconsistency}, and tree search~\citep{yao2023tree} scale test-time computation through explicit text, and allocating that computation adaptively can outperform scaling model size~\citep{snell2025scaling}.
Latent reasoning moves this computation into continuous space~\citep{zhu2025survey}.
Training-based approaches add pause tokens~\citep{goyal2024pause}, distill explicit chains into hidden states~\citep{deng2023implicit,shen2025codi}, feed hidden states back as continuous thoughts~\citep{hao2024coconut}, generate soft thoughts with an assistant model~\citep{xu2025softcot}, iterate a recurrent block in depth~\citep{geiping2025recurrent}, or scale and optimize latent thoughts at test time~\citep{you2025parallel,du2025lto}.
Training-free approaches replace sampled tokens with mixtures of embeddings~\citep{zhang2025softthinking} or switch between latent and explicit thinking according to the model's confidence~\citep{shi2025swireasoning}.
Our method organizes inference as alternating global and local belief updates and stops when the collective state stabilizes, shifting the focus from scaling computation to shaping its dynamics.

\textbf{Belief and opinion dynamics in collective computation.}
Classical models of social influence study how locally interacting opinions evolve toward consensus, polarization, clustering, or persistent disagreement, depending on the update rule and interaction graph \citep{degroot1974reaching,friedkin1990social,hegselmann2002opinion,deffuant2000mixing,castellano2009statistical}, and when such interactions aggregate dispersed information or amplify errors \citep{golub2010naive,acemoglu2011opinion,jadbabaie2012nonbayesian}. Related ideas appear in distributed computation, where belief propagation, consensus protocols, and collective dynamical systems compute global quantities through repeated local updates and relaxation toward stable states \citep{pearl1988probabilistic,yedidia2005constructing,tsitsiklis1986distributed,boyd2006randomized,hopfield1982neural}. Recent work has begun to observe analogous phenomena in populations of language-model agents interacting through text, including consensus, conformity, shared conventions, and collective biases \citep{chuang2024simulating,zhang2024exploring,estornell2024multi,ashery2025emergent,cisneros2025biases}. We extend these questions to agents whose states are high-dimensional latent beliefs updated by a frozen language model through nonlinear dynamics. Instead of assuming classical convergence guarantees, we empirically study whether these beliefs settle, how interaction reshapes the collective state, and whether equilibrium requires agreement.

% Importantly, equilibrium need not imply consensus: heterogeneous or clustered configurations can themselves be stable fixed points \citep{abelson1964mathematical,deffuant2000mixing,altafini2013consensus,bindel2015how}.

\section{Conclusion}

We introduced \emph{Stream of Belief} (SoB), formulating multi-agent reasoning as the evolution of interacting latent beliefs toward a stable joint state. By coupling global interaction with local refinement, SoB enables agents to collaborate through high-dimensional semantic spaces via belief tokens. Across diverse reasoning tasks and model families, our experiments demonstrate the effectiveness and efficiency of this approach, while revealing spontaneous alignment between evolving beliefs and the model's native token representations. More broadly, our findings suggest that collective intelligence can emerge through latent belief dynamics, offering a complementary perspective on multi-agent reasoning beyond language-based communication.

\textbf{Broad Impact.}
SoB opens a direction for studying collective intelligence as a dynamical process in continuous representation spaces. This perspective may inform more efficient and expressive multi-agent systems, while motivating further investigation into how information propagates, beliefs interact, and collective behavior emerges in language models.

\textbf{Limitations.}
Our framework relies on frozen language models and fixed-length belief representations, which may constrain the complexity of information exchanged among agents. Moreover, stable decoded predictions do not guarantee convergence to a latent fixed point or correctness of the collective solution. Finally, while latent communication reduces sequential generation and explicit text usage, it introduces computational overhead and makes intermediate reasoning less directly interpretable than natural-language communication.

\section*{Acknowledgments}
This work is supported by the Defense Advanced Research Projects Agency (DARPA) under the Young Faculty Award (YFA) program. The views and conclusions are those of the authors and should not be interpreted as representing the official policies of DARPA or the U.S. Government.

\bibliographystyle{abbrvnat}
\bibliography{references}

@article{wei2022chain,
  title={Chain of thought prompting elicits reasoning in large language models},
  author={Wei, Jason and Wang, Xuezhi and Schuurmans, Dale and Bosma, Maarten and Chi, Ed and Le, Quoc and Zhou, Denny},
  journal={arXiv preprint arXiv:2201.11903},
  year={2022}
}

@inproceedings{snell2025scaling,
  title={Scaling {LLM} Test-Time Compute Optimally Can be More Effective than Scaling Parameters for Reasoning},
  author={Snell, Charlie and Lee, Jaehoon and Xu, Kelvin and Kumar, Aviral},
  booktitle={International Conference on Learning Representations},
  year={2025}
}

@article{degroot1974reaching,
  title={Reaching a Consensus},
  author={DeGroot, Morris H.},
  journal={Journal of the American Statistical Association},
  volume={69},
  number={345},
  pages={118--121},
  year={1974},
  doi={10.1080/01621459.1974.10480137}
}

@article{friedkin1990social,
  title={Social Influence and Opinions},
  author={Friedkin, Noah E. and Johnsen, Eugene C.},
  journal={The Journal of Mathematical Sociology},
  volume={15},
  number={3-4},
  pages={193--206},
  year={1990},
  doi={10.1080/0022250X.1990.9990069}
}

@article{golub2010naive,
  title={Naive Learning in Social Networks and the Wisdom of Crowds},
  author={Golub, Benjamin and Jackson, Matthew O.},
  journal={American Economic Journal: Microeconomics},
  volume={2},
  number={1},
  pages={112--149},
  year={2010},
  doi={10.1257/mic.2.1.112}
}

@article{hopfield1982neural,
  title   = {Neural Networks and Physical Systems with Emergent Collective Computational Abilities},
  author  = {Hopfield, J. J.},
  journal = {Proceedings of the National Academy of Sciences},
  volume  = {79},
  number  = {8},
  pages   = {2554--2558},
  year    = {1982},
  doi     = {10.1073/pnas.79.8.2554}
}

@inproceedings{hao2024coconut,
  title     = {Training Large Language Models to Reason in a Continuous Latent Space},
  author    = {Hao, Shibo and Sukhbaatar, Sainbayar and Su, DiJia and Li, Xian and Hu, Zhiting and Weston, Jason and Tian, Yuandong},
  booktitle = {Conference on Language Modeling},
  year      = {2025}
}

@inproceedings{zhang2025softthinking,
  title     = {Soft Thinking: Unlocking the Reasoning Potential of {LLMs} in Continuous Concept Space},
  author    = {Zhang, Zhen and He, Xuehai and Yan, Weixiang and Shen, Ao and Zhao, Chenyang and Wang, Xin Eric},
  booktitle = {Advances in Neural Information Processing Systems},
  volume    = {38},
  year      = {2025}
}

@inproceedings{shi2025swireasoning,
  title     = {{SwiReasoning}: Switch-Thinking in Latent and Explicit for Pareto-Superior Reasoning {LLMs}},
  author    = {Shi, Dachuan and Asi, Abedelkadir and Li, Keying and Yuan, Xiangchi and Pan, Leyan and Lee, Wenke and Xiao, Wen},
  booktitle = {International Conference on Learning Representations},
  year      = {2026}
}

@inproceedings{xu2025softcot,
  title     = {{SoftCoT}: Soft Chain-of-Thought for Efficient Reasoning with {LLMs}},
  author    = {Xu, Yige and Guo, Xu and Zeng, Zhiwei and Miao, Chunyan},
  booktitle = {Proceedings of the 63rd Annual Meeting of the Association for Computational Linguistics (Volume 1: Long Papers)},
  pages     = {23336--23351},
  publisher = {Association for Computational Linguistics},
  year      = {2025},
  doi       = {10.18653/v1/2025.acl-long.1137}
}

@inproceedings{shen2025codi,
  title     = {{CODI}: Compressing Chain-of-Thought into Continuous Space via Self-Distillation},
  author    = {Shen, Zhenyi and Yan, Hanqi and Zhang, Linhai and Hu, Zhanghao and Du, Yali and He, Yulan},
  booktitle = {Proceedings of the 2025 Conference on Empirical Methods in Natural Language Processing},
  pages     = {677--693},
  publisher = {Association for Computational Linguistics},
  year      = {2025},
  doi       = {10.18653/v1/2025.emnlp-main.36}
}

@inproceedings{goyal2024pause,
  title     = {Think before you speak: Training Language Models With Pause Tokens},
  author    = {Goyal, Sachin and Ji, Ziwei and Rawat, Ankit Singh and Menon, Aditya Krishna and Kumar, Sanjiv and Nagarajan, Vaishnavh},
  booktitle = {International Conference on Learning Representations},
  year      = {2024}
}

@inproceedings{geiping2025recurrent,
  title     = {Scaling up Test-Time Compute with Latent Reasoning: A Recurrent Depth Approach},
  author    = {Geiping, Jonas and McLeish, Sean and Jain, Neel and Kirchenbauer, John and Singh, Siddharth and Bartoldson, Brian R. and Kailkhura, Bhavya and Bhatele, Abhinav and Goldstein, Tom},
  booktitle = {Advances in Neural Information Processing Systems},
  volume    = {38},
  year      = {2025}
}

@misc{zhu2025survey,
  title         = {A Survey on Latent Reasoning},
  author        = {Zhu, Rui-Jie and Peng, Tianhao and Cheng, Tianhao and Qu, Xingwei and Huang, Jinfa and Zhu, Dawei and Wang, Hao and Xue, Kaiwen and Zhang, Xuanliang and Shan, Yong and Cai, Tianle and Kergan, Taylor and Kembay, Assel and Smith, Andrew and Lin, Chenghua and others},
  year          = {2025},
  eprint        = {2507.06203},
  archivePrefix = {arXiv},
  primaryClass  = {cs.CL}
}

@misc{deng2023implicit,
  title         = {Implicit Chain of Thought Reasoning via Knowledge Distillation},
  author        = {Deng, Yuntian and Prasad, Kiran and Fernandez, Roland and Smolensky, Paul and Chaudhary, Vishrav and Shieber, Stuart},
  year          = {2023},
  eprint        = {2311.01460},
  archivePrefix = {arXiv},
  primaryClass  = {cs.CL}
}

@inproceedings{you2025parallel,
  title     = {Parallel Test-Time Scaling for Latent Reasoning Models},
  author    = {You, Runyang and Li, Yongqi and Liu, Meng and Wang, Wenjie and Nie, Liqiang and Li, Wenjie},
  booktitle = {Proceedings of the 64th Annual Meeting of the Association for Computational Linguistics (Volume 1: Long Papers)},
  pages     = {44703--44717},
  publisher = {Association for Computational Linguistics},
  year      = {2026},
  doi       = {10.18653/v1/2026.acl-long.2069}
}

@inproceedings{du2025lto,
  title     = {Latent Thinking Optimization: Your Latent Reasoning Language Model Secretly Encodes Reward Signals in Its Latent Thoughts},
  author    = {Du, Hanwen and Dong, Yuxin and Ning, Xia},
  booktitle = {International Conference on Learning Representations},
  year      = {2026}
}

@inproceedings{zou2025latentmas,
  title     = {Latent Collaboration in Multi-Agent Systems},
  author    = {Zou, Jiaru and Qiu, Ruizhong and Li, Gaotang and Yang, Xiyuan and Tieu, Katherine and Lu, Pan and Shen, Ke and Tong, Hanghang and Choi, Yejin and He, Jingrui and Zou, James and Wang, Mengdi and Yang, Ling},
  booktitle = {International Conference on Machine Learning},
  year      = {2026}
}

@inproceedings{fu2025c2c,
  title     = {Cache-to-Cache: Direct Semantic Communication Between Large Language Models},
  author    = {Fu, Tianyu and Min, Zihan and Zhang, Hanling and Yan, Jichao and Dai, Guohao and Ouyang, Wanli and Wang, Yu},
  booktitle = {International Conference on Learning Representations},
  year      = {2026}
}

@inproceedings{zheng2025thoughtcomm,
  title     = {Thought Communication in Multiagent Collaboration},
  author    = {Zheng, Yujia and Zhao, Zhuokai and Li, Zijian and Xie, Yaqi and Gao, Mingze and Zhang, Lizhu and Zhang, Kun},
  booktitle = {Advances in Neural Information Processing Systems},
  volume    = {38},
  year      = {2025}
}

@inproceedings{du2025interlat,
  title     = {Enabling Agents to Communicate Entirely in Latent Space},
  author    = {Du, Zhuoyun and Wang, Runze and Bai, Huiyu and Cao, Zouying and Zhu, Xiaoyong and Cheng, Yu and Zheng, Bo and Chen, Wei and Ying, Haochao},
  booktitle = {Proceedings of the 64th Annual Meeting of the Association for Computational Linguistics (Volume 1: Long Papers)},
  pages     = {27106--27129},
  publisher = {Association for Computational Linguistics},
  year      = {2026},
  doi       = {10.18653/v1/2026.acl-long.1248}
}

@inproceedings{ramesh2025activations,
  title     = {Communicating Activations Between Language Model Agents},
  author    = {Ramesh, Vignav and Li, Kenneth},
  booktitle = {Proceedings of the 42nd International Conference on Machine Learning},
  series    = {Proceedings of Machine Learning Research},
  volume    = {267},
  pages     = {51094--51116},
  publisher = {PMLR},
  year      = {2025}
}

@inproceedings{shi2025kvcomm,
  title     = {{KVComm}: Enabling Efficient {LLM} Communication through Selective {KV} Sharing},
  author    = {Shi, Xiangyu and Chiesa, Marco and Maguire Jr., Gerald Q. and Kostic, Dejan},
  booktitle = {International Conference on Learning Representations},
  year      = {2026}
}

@inproceedings{tang2025sde,
  title     = {Augmenting Multi-Agent Communication with State Delta Trajectory},
  author    = {Tang, Yichen and Su, Weihang and Zhou, Yujia and Liu, Yiqun and Zhang, Min and Ma, Shaoping and Ai, Qingyao},
  booktitle = {Proceedings of the 2025 Conference on Empirical Methods in Natural Language Processing},
  pages     = {10219--10240},
  publisher = {Association for Computational Linguistics},
  year      = {2025},
  doi       = {10.18653/v1/2025.emnlp-main.518}
}

@inproceedings{du2024debate,
  title     = {Improving Factuality and Reasoning in Language Models through Multiagent Debate},
  author    = {Du, Yilun and Li, Shuang and Torralba, Antonio and Tenenbaum, Joshua B. and Mordatch, Igor},
  booktitle = {Proceedings of the 41st International Conference on Machine Learning},
  series    = {Proceedings of Machine Learning Research},
  volume    = {235},
  pages     = {11733--11763},
  year      = {2024}
}

@inproceedings{liang2024divergent,
  title     = {Encouraging Divergent Thinking in Large Language Models through Multi-Agent Debate},
  author    = {Liang, Tian and He, Zhiwei and Jiao, Wenxiang and Wang, Xing and Wang, Yan and Wang, Rui and Yang, Yujiu and Shi, Shuming and Tu, Zhaopeng},
  booktitle = {Proceedings of the 2024 Conference on Empirical Methods in Natural Language Processing},
  pages     = {17889--17904},
  year      = {2024}
}

@inproceedings{chan2024chateval,
  title     = {{ChatEval}: Towards Better {LLM}-based Evaluators through Multi-Agent Debate},
  author    = {Chan, Chi-Min and Chen, Weize and Su, Yusheng and Yu, Jianxuan and Xue, Wei and Zhang, Shanghang and Fu, Jie and Liu, Zhiyuan},
  booktitle = {International Conference on Learning Representations},
  year      = {2024}
}

@inproceedings{wu2024autogen,
  title     = {{AutoGen}: Enabling Next-Gen {LLM} Applications via Multi-Agent Conversation},
  author    = {Wu, Qingyun and Bansal, Gagan and Zhang, Jieyu and Wu, Yiran and Li, Beibin and Zhu, Erkang and Jiang, Li and Zhang, Xiaoyun and Zhang, Shaokun and Liu, Jiale and Awadallah, Ahmed Hassan and White, Ryen W. and Burger, Doug and Wang, Chi},
  booktitle = {Conference on Language Modeling},
  year      = {2024}
}

@inproceedings{li2023camel,
  title     = {{CAMEL}: Communicative Agents for ``Mind'' Exploration of Large Language Model Society},
  author    = {Li, Guohao and Hammoud, Hasan Abed Al Kader and Itani, Hani and Khizbullin, Dmitrii and Ghanem, Bernard},
  booktitle = {Advances in Neural Information Processing Systems},
  volume    = {36},
  year      = {2023}
}

@inproceedings{hong2024metagpt,
  title     = {{MetaGPT}: Meta Programming for A Multi-Agent Collaborative Framework},
  author    = {Hong, Sirui and Zhuge, Mingchen and Chen, Jonathan and Zheng, Xiawu and Cheng, Yuheng and Wang, Jinlin and Zhang, Ceyao and Wang, Zili and Yau, Steven Ka Shing and Lin, Zijuan and Zhou, Liyang and Ran, Chenyu and Xiao, Lingfeng and Wu, Chenglin and Schmidhuber, J{\"u}rgen},
  booktitle = {International Conference on Learning Representations},
  year      = {2024}
}

@inproceedings{park2023generative,
  title     = {Generative Agents: Interactive Simulacra of Human Behavior},
  author    = {Park, Joon Sung and O'Brien, Joseph C. and Cai, Carrie J. and Morris, Meredith Ringel and Liang, Percy and Bernstein, Michael S.},
  booktitle = {Proceedings of the 36th Annual ACM Symposium on User Interface Software and Technology},
  doi       = {10.1145/3586183.3606763},
  year      = {2023}
}

@inproceedings{zhuge2024gptswarm,
  title     = {{GPTSwarm}: Language Agents as Optimizable Graphs},
  author    = {Zhuge, Mingchen and Wang, Wenyi and Kirsch, Louis and Faccio, Francesco and Khizbullin, Dmitrii and Schmidhuber, J{\"u}rgen},
  booktitle = {Proceedings of the 41st International Conference on Machine Learning},
  series    = {Proceedings of Machine Learning Research},
  volume    = {235},
  pages     = {62743--62767},
  year      = {2024}
}

@inproceedings{qian2025scaling,
  title     = {Scaling Large Language Model-based Multi-Agent Collaboration},
  author    = {Qian, Chen and Xie, Zihao and Wang, YiFei and Liu, Wei and Zhu, Kunlun and Xia, Hanchen and Dang, Yufan and Du, Zhuoyun and Chen, Weize and Yang, Cheng and Liu, Zhiyuan and Sun, Maosong},
  booktitle = {International Conference on Learning Representations},
  year      = {2025}
}

@article{li2024moreagents,
  title   = {More Agents Is All You Need},
  author  = {Li, Junyou and Zhang, Qin and Yu, Yangbin and Fu, Qiang and Ye, Deheng},
  journal = {Transactions on Machine Learning Research},
  year    = {2024}
}

@inproceedings{zhang2025gdesigner,
  title     = {{G-Designer}: Architecting Multi-agent Communication Topologies via Graph Neural Networks},
  author    = {Zhang, Guibin and Yue, Yanwei and Sun, Xiangguo and Wan, Guancheng and Yu, Miao and Fang, Junfeng and Wang, Kun and Chen, Tianlong and Cheng, Dawei},
  booktitle = {Proceedings of the 42nd International Conference on Machine Learning},
  series    = {Proceedings of Machine Learning Research},
  volume    = {267},
  pages     = {76678--76692},
  year      = {2025}
}

@inproceedings{chen2024agentverse,
  title     = {{AgentVerse}: Facilitating Multi-Agent Collaboration and Exploring Emergent Behaviors},
  author    = {Chen, Weize and Su, Yusheng and Zuo, Jingwei and Yang, Cheng and Yuan, Chenfei and Chan, Chi-Min and Yu, Heyang and Lu, Yaxi and Hung, Yi-Hsin and Qian, Chen and Qin, Yujia and Cong, Xin and Xie, Ruobing and Liu, Zhiyuan and Sun, Maosong and Zhou, Jie},
  booktitle = {International Conference on Learning Representations},
  year      = {2024}
}

@inproceedings{wang2023selfconsistency,
  title     = {Self-Consistency Improves Chain of Thought Reasoning in Language Models},
  author    = {Wang, Xuezhi and Wei, Jason and Schuurmans, Dale and Le, Quoc and Chi, Ed and Narang, Sharan and Chowdhery, Aakanksha and Zhou, Denny},
  booktitle = {International Conference on Learning Representations},
  year      = {2023}
}

@inproceedings{yao2023tree,
  title     = {Tree of Thoughts: Deliberate Problem Solving with Large Language Models},
  author    = {Yao, Shunyu and Yu, Dian and Zhao, Jeffrey and Shafran, Izhak and Griffiths, Thomas L. and Cao, Yuan and Narasimhan, Karthik},
  booktitle = {Advances in Neural Information Processing Systems},
  volume    = {36},
  pages     = {11809--11822},
  year      = {2023}
}

@misc{cobbe2021gsm8k,
  title         = {Training Verifiers to Solve Math Word Problems},
  author        = {Cobbe, Karl and Kosaraju, Vineet and Bavarian, Mohammad and Chen, Mark and Jun, Heewoo and Kaiser, Lukasz and Plappert, Matthias and Tworek, Jerry and Hilton, Jacob and Nakano, Reiichiro and Hesse, Christopher and Schulman, John},
  year          = {2021},
  eprint        = {2110.14168},
  archivePrefix = {arXiv}
}

@inproceedings{hendrycks2021math,
  title     = {Measuring Mathematical Problem Solving With the {MATH} Dataset},
  author    = {Hendrycks, Dan and Burns, Collin and Kadavath, Saurav and Arora, Akul and Basart, Steven and Tang, Eric and Song, Dawn and Steinhardt, Jacob},
  booktitle = {Proceedings of the Neural Information Processing Systems Track on Datasets and Benchmarks},
  volume    = {1},
  year      = {2021}
}

@inproceedings{lightman2024verify,
  title     = {Let's Verify Step by Step},
  author    = {Lightman, Hunter and Kosaraju, Vineet and Burda, Yuri and Edwards, Harrison and Baker, Bowen and Lee, Teddy and Leike, Jan and Schulman, John and Sutskever, Ilya and Cobbe, Karl},
  booktitle = {International Conference on Learning Representations},
  year      = {2024}
}

@inproceedings{rein2024gpqa,
  title     = {{GPQA}: A Graduate-Level {Google}-Proof {Q\&A} Benchmark},
  author    = {Rein, David and Hou, Betty Li and Stickland, Asa Cooper and Petty, Jackson and Pang, Richard Yuanzhe and Dirani, Julien and Michael, Julian and Bowman, Samuel R.},
  booktitle = {Conference on Language Modeling},
  year      = {2024}
}

@misc{clark2018arc,
  title         = {Think you have Solved Question Answering? Try {ARC}, the {AI2} Reasoning Challenge},
  author        = {Clark, Peter and Cowhey, Isaac and Etzioni, Oren and Khot, Tushar and Sabharwal, Ashish and Schoenick, Carissa and Tafjord, Oyvind},
  year          = {2018},
  eprint        = {1803.05457},
  archivePrefix = {arXiv}
}

@article{jin2021medqa,
  title   = {What Disease Does This Patient Have? A Large-Scale Open Domain Question Answering Dataset from Medical Exams},
  author  = {Jin, Di and Pan, Eileen and Oufattole, Nassim and Weng, Wei-Hung and Fang, Hanyi and Szolovits, Peter},
  journal = {Applied Sciences},
  volume  = {11},
  number  = {14},
  pages   = {6421},
  year    = {2021},
  doi     = {10.3390/app11146421}
}

@inproceedings{suzgun2023bbh,
  title     = {Challenging {BIG}-Bench Tasks and Whether Chain-of-Thought Can Solve Them},
  author    = {Suzgun, Mirac and Scales, Nathan and Sch{\"a}rli, Nathanael and Gehrmann, Sebastian and Tay, Yi and Chung, Hyung Won and Chowdhery, Aakanksha and Le, Quoc and Chi, Ed and Zhou, Denny and Wei, Jason},
  booktitle = {Findings of the Association for Computational Linguistics: ACL 2023},
  pages     = {13003--13051},
  publisher = {Association for Computational Linguistics},
  year      = {2023},
  doi       = {10.18653/v1/2023.findings-acl.824}
}

@inproceedings{yang2018hotpotqa,
  title     = {{HotpotQA}: A Dataset for Diverse, Explainable Multi-hop Question Answering},
  author    = {Yang, Zhilin and Qi, Peng and Zhang, Saizheng and Bengio, Yoshua and Cohen, William and Salakhutdinov, Ruslan and Manning, Christopher D.},
  booktitle = {Proceedings of the 2018 Conference on Empirical Methods in Natural Language Processing},
  pages     = {2369--2380},
  publisher = {Association for Computational Linguistics},
  year      = {2018},
  doi       = {10.18653/v1/D18-1259}
}

@misc{grattafiori2024llama3,
  title         = {The {Llama} 3 Herd of Models},
  author        = {Grattafiori, Aaron and Dubey, Abhimanyu and Jauhri, Abhinav and Pandey, Abhinav and Kadian, Abhishek and others},
  year          = {2024},
  eprint        = {2407.21783},
  archivePrefix = {arXiv}
}

@misc{gemma2025gemma3,
  title         = {{Gemma} 3 Technical Report},
  author        = {{Gemma Team}},
  year          = {2025},
  eprint        = {2503.19786},
  archivePrefix = {arXiv}
}

@misc{abouelenin2025phi4mini,
  title         = {{Phi-4-Mini} Technical Report: Compact yet Powerful Multimodal Language Models via Mixture-of-{LoRAs}},
  author        = {Abouelenin, Abdelrahman and Ashfaq, Atabak and Atkinson, Adam and Awadalla, Hany and Bach, Nguyen and Bao, Jianmin and others},
  year          = {2025},
  eprint        = {2503.01743},
  archivePrefix = {arXiv}
}

@article{woolley2010evidence,
  author  = {Woolley, Anita Williams and Chabris, Christopher F. and Pentland, Alex and Hashmi, Nada and Malone, Thomas W.},
  title   = {Evidence for a collective intelligence factor in the performance of human groups},
  journal = {Science},
  year    = {2010},
  volume  = {330},
  number  = {6004},
  pages   = {686--688},
  doi     = {10.1126/science.1193147}
}

@article{hong2004groups,
  author  = {Hong, Lu and Page, Scott E.},
  title   = {Groups of diverse problem solvers can outperform groups of high-ability problem solvers},
  journal = {Proceedings of the National Academy of Sciences},
  year    = {2004},
  volume  = {101},
  number  = {46},
  pages   = {16385--16389},
  doi     = {10.1073/pnas.0403723101}
}

@article{lorenz2011social,
  author  = {Lorenz, Jan and Rauhut, Heiko and Schweitzer, Frank and Helbing, Dirk},
  title   = {How social influence can undermine the wisdom of crowd effect},
  journal = {Proceedings of the National Academy of Sciences},
  year    = {2011},
  volume  = {108},
  number  = {22},
  pages   = {9020--9025},
  doi     = {10.1073/pnas.1008636108}
}

@article{galton1907vox,
  author  = {Galton, Francis},
  title   = {Vox populi},
  journal = {Nature},
  year    = {1907},
  volume  = {75},
  number  = {1949},
  pages   = {450--451},
  doi     = {10.1038/075450a0}
}

@article{couzin2009collective,
  author  = {Couzin, Iain D.},
  title   = {Collective cognition in animal groups},
  journal = {Trends in Cognitive Sciences},
  year    = {2009},
  volume  = {13},
  number  = {1},
  pages   = {36--43},
  doi     = {10.1016/j.tics.2008.10.002}
}

@article{hegselmann2002opinion,
  author  = {Hegselmann, Rainer and Krause, Ulrich},
  title   = {Opinion dynamics and bounded confidence: Models, analysis and simulation},
  journal = {Journal of Artificial Societies and Social Simulation},
  year    = {2002},
  volume  = {5},
  number  = {3},
  pages   = {2}
}

@inproceedings{zhang2024chain,
  title     = {Chain of Agents: Large Language Models Collaborating on Long-Context Tasks},
  author    = {Zhang, Yusen and Sun, Ruoxi and Chen, Yanfei and Pfister, Tomas and Zhang, Rui and Arik, Sercan},
  booktitle = {Advances in Neural Information Processing Systems},
  volume    = {37},
  pages     = {132208--132237},
  year      = {2024}
}

@inproceedings{zhao2025sirius,
  title     = {{SiriuS}: Self-improving Multi-agent Systems via Bootstrapped Reasoning},
  author    = {Zhao, Wanjia and Yuksekgonul, Mert and Wu, Shirley and Zou, James},
  booktitle = {Advances in Neural Information Processing Systems},
  volume    = {38},
  year      = {2025}
}

@inproceedings{ye2025kvcomm,
  title     = {{KVCOMM}: Online Cross-context {KV}-cache Communication for Efficient {LLM}-based Multi-agent Systems},
  author    = {Ye, Hancheng and Gao, Zhengqi and Ma, Mingyuan and Wang, Qinsi and Fu, Yuzhe and Chung, Ming-Yu and Lin, Yueqian and Liu, Zhijian and Zhang, Jianyi and Zhuo, Danyang and others},
  booktitle = {Advances in Neural Information Processing Systems},
  volume    = {38},
  year      = {2025}
}

@article{yang2025qwen3,
  title={Qwen3 technical report},
  author={Yang, An and Li, Anfeng and Yang, Baosong and Zhang, Beichen and Hui, Binyuan and Zheng, Bo and Yu, Bowen and Gao, Chang and Huang, Chengen and Lv, Chenxu and others},
  journal={arXiv preprint arXiv:2505.09388},
  year={2025}
}

@article{tran2025multi,
  title={Multi-agent collaboration mechanisms: A survey of llms},
  author={Tran, Khanh-Tung and Dao, Dung and Nguyen, Minh-Duong and Pham, Quoc-Viet and O'Sullivan, Barry and Nguyen, Hoang D},
  journal={arXiv preprint arXiv:2501.06322},
  year={2025}
}

@inproceedings{hu2025owl,
  title     = {{OWL}: Optimized Workforce Learning for General Multi-Agent Assistance in Real-World Task Automation},
  author    = {Hu, Mengkang and Zhou, Yuhang and Fan, Wendong and Nie, Yuzhou and Xia, Bowei and Sun, Tao and Ye, Ziyu and Jin, Zhaoxuan and Li, Yingru and Chen, Qiguang and others},
  booktitle = {Advances in Neural Information Processing Systems},
  volume    = {38},
  year      = {2025}
}

@inproceedings{guo2024large,
  title     = {Large Language Model Based Multi-Agents: A Survey of Progress and Challenges},
  author    = {Guo, Taicheng and Chen, Xiuying and Wang, Yaqi and Chang, Ruidi and Pei, Shichao and Chawla, Nitesh V. and Wiest, Olaf and Zhang, Xiangliang},
  booktitle = {Proceedings of the Thirty-Third International Joint Conference on Artificial Intelligence (IJCAI-24)},
  pages     = {8048--8057},
  year      = {2024},
  doi       = {10.24963/ijcai.2024/890}
}

@article{coda2025exploring,
  title={Exploring System 1 and 2 communication for latent reasoning in LLMs},
  author={Coda-Forno, Julian and Zhao, Zhuokai and Zhang, Qiang and Tamboli, Dipesh and Li, Weiwei and Fan, Xiangjun and Zhang, Lizhu and Schulz, Eric and Tseng, Hsiao-Ping},
  journal={arXiv preprint arXiv:2510.00494},
  year={2025}
}

@article{acharya2025agentic,
  title   = {Agentic {AI}: Autonomous Intelligence for Complex Goals---A Comprehensive Survey},
  author  = {Acharya, Deepak Bhaskar and Kuppan, Karthigeyan and Divya, B.},
  journal = {IEEE Access},
  volume  = {13},
  pages   = {18912--18936},
  year    = {2025},
  doi     = {10.1109/ACCESS.2025.3532853}
}

@inproceedings{zhou2025reso,
  title     = {{ReSo}: A Reward-driven Self-organizing {LLM}-based Multi-Agent System for Reasoning Tasks},
  author    = {Zhou, Heng and Geng, Hejia and Xue, Xiangyuan and Kang, Li and Qin, Yiran and Wang, Zhiyong and Yin, Zhenfei and Bai, Lei},
  booktitle = {Proceedings of the 2025 Conference on Empirical Methods in Natural Language Processing},
  pages     = {15979--15998},
  year      = {2025},
  doi       = {10.18653/v1/2025.emnlp-main.808}
}

@article{li2025flow,
  title={In-the-Flow Agentic System Optimization for Effective Planning and Tool Use},
  author={Li, Zhuofeng and Zhang, Haoxiang and Han, Seungju and Liu, Sheng and Xie, Jianwen and Zhang, Yu and Choi, Yejin and Zou, James and Lu, Pan},
  journal={arXiv preprint arXiv:2510.05592},
  year={2025}
}

@article{yang2024llm,
  title={LLM-based Multi-Agent Systems: Techniques and Business Perspectives},
  author={Yang, Yingxuan and Peng, Qiuying and Wang, Jun and Wen, Ying and Zhang, Weinan},
  journal={arXiv preprint arXiv:2411.14033},
  year={2024}
}

@inproceedings{tao2024magis,
  title     = {{MAGIS}: {LLM}-Based Multi-Agent Framework for {GitHub} Issue Resolution},
  author    = {Tao, Wei and Zhou, Yucheng and Wang, Yanlin and Zhang, Wenqiang and Zhang, Hongyu and Cheng, Yu},
  booktitle = {Advances in Neural Information Processing Systems},
  volume    = {37},
  pages     = {51963--51993},
  year      = {2024}
}

@article{wang2025talk,
  title   = {Talk Structurally, Act Hierarchically: A Collaborative Framework for {LLM} Multi-Agent Systems},
  author  = {Wang, Zhao and Moriyama, Sota and Wang, Wei-Yao and Gangopadhyay, Briti and Takamatsu, Shingo},
  journal = {IEEE Transactions on Artificial Intelligence},
  volume  = {7},
  number  = {9},
  pages   = {4993--5003},
  year    = {2026},
  doi     = {10.1109/TAI.2026.3676025}
}

@inproceedings{yao2022react,
  title     = {{ReAct}: Synergizing Reasoning and Acting in Language Models},
  author    = {Yao, Shunyu and Zhao, Jeffrey and Yu, Dian and Du, Nan and Shafran, Izhak and Narasimhan, Karthik R. and Cao, Yuan},
  booktitle = {International Conference on Learning Representations},
  year      = {2023}
}

@article{mieczkowski2025predicting,
  title={Predicting Multi-Agent Specialization via Task Parallelizability},
  author={Mieczkowski, Elizabeth and Mon-Williams, Ruaridh and Bramley, Neil and Lucas, Christopher G and Velez, Natalia and Griffiths, Thomas L},
  journal={arXiv preprint arXiv:2503.15703},
  year={2025}
}

@inproceedings{
huang2025many,
title={Many Minds, One Goal: Time Series Forecasting via Sub-task Specialization and Inter-agent Cooperation},
author={Qihe Huang and Zhengyang Zhou and Yangze Li and Kuo Yang and Binwu Wang and Yang Wang},
booktitle={The Thirty-ninth Annual Conference on Neural Information Processing Systems},
year={2026},
url={https://openreview.net/forum?id=Uon41HfqR3}
}

@misc{gsm8k,
      title={Training Verifiers to Solve Math Word Problems}, 
      author={Karl Cobbe and Vineet Kosaraju and Mohammad Bavarian and Mark Chen and Heewoo Jun and Lukasz Kaiser and Matthias Plappert and Jerry Tworek and Jacob Hilton and Reiichiro Nakano and Christopher Hesse and John Schulman},
      year={2021},
      eprint={2110.14168},
      archivePrefix={arXiv},
      primaryClass={cs.LG},
      url={https://arxiv.org/abs/2110.14168}, 
}

@article{Eywa2026,
  author       = {Zihao Li and
                  Jiaru Zou and
                  Feihao Fang and
                  Xuying Ning and
                  Mengting Ai and
                  Tianxin Wei and
                  Sirui Chen and
                  Xiyuan Yang and
                  Jingrui He},
  title        = {Heterogeneous Scientific Foundation Model Collaboration},
  journal      = {CoRR},
  volume       = {abs/2604.27351},
  year         = {2026},
  url          = {https://doi.org/10.48550/arXiv.2604.27351},
  doi          = {10.48550/ARXIV.2604.27351},
  eprinttype   = {arXiv},
  eprint       = {2604.27351},
  bibsource    = {dblp computer science bibliography, https://dblp.org}
}

@article{medqa,
  title={Llm-medqa: Enhancing medical question answering through case studies in large language models},
  author={Yang, Hang and Chen, Hao and Guo, Hui and Chen, Yineng and Lin, Ching-Sheng and Hu, Shu and Hu, Jinrong and Wu, Xi and Wang, Xin},
  journal={arXiv preprint arXiv:2501.05464},
  year={2024}
}

@String(ACL = "{Annual Meeting of the Association for Computational Linguistics (ACL)}")

@String(NAACL = "{Annual Conference of the North American Chapter of the Association for Computational Linguistics (NAACL)}")

@inproceedings{wang2024mmlupro,
  title={MMLU-Pro: A More Robust and Challenging Multi-Task Language Understanding Benchmark},
  author={Wang, Yubo and Ma, Xueguang and Zhang, Ge and Ni, Yuansheng and Chandra, Abhranil and Guo, Shiguang and Ren, Weiming and Arulraj, Aaran and He, Xuan and Jiang, Ziyan and Li, Tianle and Ku, Max and Wang, Kai and Zhuang, Alex and Fan, Rongqi and Yue, Xiang and Chen, Wenhu},
  booktitle={Advances in Neural Information Processing Systems},
  year={2024}
}

@incollection{abelson1964mathematical,
  title     = {Mathematical Models of the Distribution of Attitudes under Controversy},
  author    = {Abelson, Robert P.},
  booktitle = {Contributions to Mathematical Psychology},
  editor    = {Frederiksen, Norman and Gulliksen, Harold},
  publisher = {Holt, Rinehart and Winston},
  address   = {New York},
  pages     = {142--160},
  year      = {1964}
}

@article{deffuant2000mixing,
  title   = {Mixing Beliefs among Interacting Agents},
  author  = {Deffuant, Guillaume and Neau, David and Amblard, Fr{\'e}d{\'e}ric and Weisbuch, G{\'e}rard},
  journal = {Advances in Complex Systems},
  volume  = {3},
  number  = {1--4},
  pages   = {87--98},
  year    = {2000},
  doi     = {10.1142/S0219525900000078}
}

@article{castellano2009statistical,
  title   = {Statistical Physics of Social Dynamics},
  author  = {Castellano, Claudio and Fortunato, Santo and Loreto, Vittorio},
  journal = {Reviews of Modern Physics},
  volume  = {81},
  number  = {2},
  pages   = {591--646},
  year    = {2009},
  doi     = {10.1103/RevModPhys.81.591}
}

@article{acemoglu2011opinion,
  title   = {Opinion Dynamics and Learning in Social Networks},
  author  = {Acemoglu, Daron and Ozdaglar, Asuman},
  journal = {Dynamic Games and Applications},
  volume  = {1},
  number  = {1},
  pages   = {3--49},
  year    = {2011},
  doi     = {10.1007/s13235-010-0004-1}
}

@article{acemoglu2013opinion,
  title   = {Opinion Fluctuations and Disagreement in Social Networks},
  author  = {Acemoglu, Daron and Como, Giacomo and Fagnani, Fabio and Ozdaglar, Asuman},
  journal = {Mathematics of Operations Research},
  volume  = {38},
  number  = {1},
  pages   = {1--27},
  year    = {2013},
  doi     = {10.1287/moor.1120.0570}
}

@article{bindel2015how,
  title   = {How Bad Is Forming Your Own Opinion?},
  author  = {Bindel, David and Kleinberg, Jon and Oren, Sigal},
  journal = {Games and Economic Behavior},
  volume  = {92},
  pages   = {248--265},
  year    = {2015},
  doi     = {10.1016/j.geb.2014.06.004}
}

@book{surowiecki2004wisdom,
  title     = {The Wisdom of Crowds: Why the Many Are Smarter Than the Few and How Collective Wisdom Shapes Business, Economies, Societies and Nations},
  author    = {Surowiecki, James},
  publisher = {Doubleday},
  address   = {New York},
  year      = {2004}
}

@article{becker2017network,
  title   = {Network Dynamics of Social Influence in the Wisdom of Crowds},
  author  = {Becker, Joshua and Brackbill, Devon and Centola, Damon},
  journal = {Proceedings of the National Academy of Sciences},
  volume  = {114},
  number  = {26},
  pages   = {E5070--E5076},
  year    = {2017},
  doi     = {10.1073/pnas.1615978114}
}

@article{jadbabaie2012nonbayesian,
  title   = {Non-{B}ayesian Social Learning},
  author  = {Jadbabaie, Ali and Molavi, Pooya and Sandroni, Alvaro and Tahbaz-Salehi, Alireza},
  journal = {Games and Economic Behavior},
  volume  = {76},
  number  = {1},
  pages   = {210--225},
  year    = {2012},
  doi     = {10.1016/j.geb.2012.06.001}
}

@book{pearl1988probabilistic,
  title     = {Probabilistic Reasoning in Intelligent Systems: Networks of Plausible Inference},
  author    = {Pearl, Judea},
  publisher = {Morgan Kaufmann},
  address   = {San Mateo, CA},
  year      = {1988}
}

@article{yedidia2005constructing,
  title   = {Constructing Free-Energy Approximations and Generalized Belief Propagation Algorithms},
  author  = {Yedidia, Jonathan S. and Freeman, William T. and Weiss, Yair},
  journal = {IEEE Transactions on Information Theory},
  volume  = {51},
  number  = {7},
  pages   = {2282--2312},
  year    = {2005},
  doi     = {10.1109/TIT.2005.850085}
}

@article{tsitsiklis1986distributed,
  title   = {Distributed Asynchronous Deterministic and Stochastic Gradient Optimization Algorithms},
  author  = {Tsitsiklis, John N. and Bertsekas, Dimitri P. and Athans, Michael},
  journal = {IEEE Transactions on Automatic Control},
  volume  = {31},
  number  = {9},
  pages   = {803--812},
  year    = {1986},
  doi     = {10.1109/TAC.1986.1104412}
}

@article{boyd2006randomized,
  title   = {Randomized Gossip Algorithms},
  author  = {Boyd, Stephen and Ghosh, Arpita and Prabhakar, Balaji and Shah, Devavrat},
  journal = {IEEE Transactions on Information Theory},
  volume  = {52},
  number  = {6},
  pages   = {2508--2530},
  year    = {2006},
  doi     = {10.1109/TIT.2006.874516}
}

@inproceedings{zhang2024exploring,
  title     = {Exploring Collaboration Mechanisms for {LLM} Agents: A Social Psychology View},
  author    = {Zhang, Jintian and Xu, Xin and Zhang, Ningyu and Liu, Ruibo and Hooi, Bryan and Deng, Shumin},
  booktitle = {Proceedings of the 62nd Annual Meeting of the Association for Computational Linguistics (Volume 1: Long Papers)},
  pages     = {14544--14607},
  year      = {2024},
  doi       = {10.18653/v1/2024.acl-long.782}
}

@inproceedings{estornell2024multi,
  title     = {Multi-{LLM} Debate: Framework, Principals, and Interventions},
  author    = {Estornell, Andrew and Liu, Yang},
  booktitle = {Advances in Neural Information Processing Systems},
  volume    = {37},
  pages     = {28938--28964},
  year      = {2024}
}

@inproceedings{chuang2024simulating,
  title     = {Simulating Opinion Dynamics with Networks of {LLM}-based Agents},
  author    = {Chuang, Yun-Shiuan and Goyal, Agam and Harlalka, Nikunj and Suresh, Siddharth and Hawkins, Robert and Yang, Sijia and Shah, Dhavan and Hu, Junjie and Rogers, Timothy T.},
  booktitle = {Findings of the Association for Computational Linguistics: NAACL 2024},
  pages     = {3326--3346},
  year      = {2024},
  doi       = {10.18653/v1/2024.findings-naacl.211}
}

@article{ashery2025emergent,
  title   = {Emergent Social Conventions and Collective Bias in {LLM} Populations},
  author  = {Ashery, Ariel Flint and Aiello, Luca Maria and Baronchelli, Andrea},
  journal = {Science Advances},
  volume  = {11},
  number  = {20},
  pages   = {eadu9368},
  year    = {2025},
  doi     = {10.1126/sciadv.adu9368}
}

@inproceedings{cisneros2025biases,
  title     = {Biases in Opinion Dynamics in Multi-Agent Systems of Large Language Models: A Case Study on Funding Allocation},
  author    = {Cisneros-Velarde, Pedro},
  booktitle = {Findings of the Association for Computational Linguistics: NAACL 2025},
  pages     = {1889--1916},
  year      = {2025},
  doi       = {10.18653/v1/2025.findings-naacl.101}
}

@inproceedings{qian2024chatdev,
  title     = {{ChatDev}: Communicative Agents for Software Development},
  author    = {Qian, Chen and Liu, Wei and Liu, Hongzhang and Chen, Nuo and Dang, Yufan and Li, Jiahao and Yang, Cheng and Chen, Weize and Su, Yusheng and Cong, Xin and Xu, Juyuan and Li, Dahai and Liu, Zhiyuan and Sun, Maosong},
  booktitle = {Proceedings of the 62nd Annual Meeting of the Association for Computational Linguistics (Volume 1: Long Papers)},
  pages     = {15174--15186},
  year      = {2024},
  doi       = {10.18653/v1/2024.acl-long.810}
}

@inproceedings{islam2024mapcoder,
  title     = {{MapCoder}: Multi-Agent Code Generation for Competitive Problem Solving},
  author    = {Islam, Md. Ashraful and Ali, Mohammed Eunus and Parvez, Md Rizwan},
  booktitle = {Proceedings of the 62nd Annual Meeting of the Association for Computational Linguistics (Volume 1: Long Papers)},
  pages     = {4912--4944},
  year      = {2024},
  doi       = {10.18653/v1/2024.acl-long.269}
}

@article{swanson2025virtual,
  title   = {The Virtual Lab of {AI} Agents Designs New {SARS-CoV-2} Nanobodies},
  author  = {Swanson, Kyle and Wu, Wesley and Bulaong, Nash L. and Pak, John E. and Zou, James},
  journal = {Nature},
  volume  = {646},
  number  = {8085},
  pages   = {716--723},
  year    = {2025},
  doi     = {10.1038/s41586-025-09442-9}
}

@article{gottweis2026accelerating,
  title   = {Accelerating Scientific Discovery with {Co-Scientist}},
  author  = {Gottweis, Juraj and Weng, Wei-Hung and Daryin, Alexander and Tu, Tao and Sirkovic, Petar and Myaskovsky, Artiom and Glowaty, Grzegorz and Weissenberger, Felix and Orlandi, Alessio and Popovici, Dan and Palepu, Anil and Rong, Keran and others},
  journal = {Nature},
  volume  = {655},
  number  = {8122},
  pages   = {487--496},
  year    = {2026},
  doi     = {10.1038/s41586-026-10644-y}
}

@inproceedings{chen2024reconcile,
  title     = {{ReConcile}: Round-Table Conference Improves Reasoning via Consensus among Diverse {LLMs}},
  author    = {Chen, Justin and Saha, Swarnadeep and Bansal, Mohit},
  booktitle = {Proceedings of the 62nd Annual Meeting of the Association for Computational Linguistics (Volume 1: Long Papers)},
  pages     = {7066--7085},
  year      = {2024},
  doi       = {10.18653/v1/2024.acl-long.381}
}

@article{szekely2013energy,
  title={Energy statistics: A class of statistics based on distances},
  author={Sz{\'e}kely, G{\'a}bor J. and Rizzo, Maria L.},
  journal={Journal of Statistical Planning and Inference},
  volume={143},
  number={8},
  pages={1249--1272},
  year={2013},
  doi={10.1016/j.jspi.2013.03.018}
}

\clearpage
\newpage
\appendix
\part*{Appendices}

\section{Prompt Templates of Stream of Belief}
\label{app:prompts}

\newtcolorbox{promptbox}[2][]{colback=black!2, colframe=softblue, colbacktitle=softblue, coltitle=white, fonttitle=\bfseries, boxrule=1.5pt, arc=2pt, left=8pt, right=8pt, top=6pt, bottom=6pt, title={#2}, #1}
\newcommand{\lat}[1]{\textcolor{softblue}{$\langle #1\rangle$}}
\newcommand{\role}[1]{\textbf{#1}}

Every belief update and every decoding step is one chat-formatted call of the frozen model (system turn, user turn, assistant turn in the model's own chat template), identical for all models and benchmarks. \lat{z_i} denotes the belief of agent $i$, inserted as $L$ soft input embeddings; \texttt{\{problem\}} is the problem text, with multiple-choice options and listed facts shown in an agent-specific random order.

\begin{promptbox}{System turn (all calls)}
\small
You are Agent $i$ in a team of $N$ agents solving a reasoning problem together. \texttt{\{persona\}}
\tcblower
\small
\textbf{Personas} (agent $i$ receives persona $i \bmod 8$):
\begin{enumerate}[nosep, leftmargin=1.5em]
\item You reason carefully and check each step.
\item You look for the most direct chain of evidence.
\item You are skeptical and try to rule out wrong options.
\item You organise the facts systematically before concluding.
\item You focus on tracing relationships one hop at a time.
\item You look for contradictions and edge cases.
\item You summarise what is certain before deciding.
\item You verify the answer by working backwards from the options.
\end{enumerate}
\end{promptbox}

\begin{promptbox}{Initial belief ($k=0$): $z_i \leftarrow A_i(x)$}
\small
\role{User:} \texttt{\{problem\}}\\
Concisely work through the key reasoning steps and end with your conclusion.\\[3pt]
\role{Assistant:} \lat{z_i} \quad ($L$ belief tokens, Eq.~\ref{eq:belief_token})
\end{promptbox}

\begin{promptbox}{Global interaction: $[\mathsf G(Z)]_i = A_i(x, z_i, \{z_j\}_{j\neq i})$}
\small
\role{User:} \texttt{\{problem\}}\\
Conclusions shared by other agents (in their own latent words):\\
{[Agent $j$]}: \lat{z_j} \quad (one line per other agent $j \neq i$)\\
Your previous conclusion: \lat{z_i}\\
Reconsider the problem and the conclusions above. Concisely work through the key reasoning steps and end with your revised conclusion.\\[3pt]
\role{Assistant:} \lat{z_i'} \quad (new belief, replaces \lat{z_i})
\end{promptbox}

\begin{promptbox}{Local refinement: $[\mathsf L(Z)]_i = A_i(x, z_i)$}
\small
\role{User:} \texttt{\{problem\}}\\
Your previous conclusion: \lat{z_i}\\
Reconsider the problem and your previous conclusion. Concisely work through the key reasoning steps and end with your revised conclusion.\\[3pt]
\role{Assistant:} \lat{z_i'} \quad (new belief, replaces \lat{z_i})
\end{promptbox}

\begin{promptbox}{Probe (after each cycle; 24 text tokens)}
\small
\role{User:} \texttt{\{problem\}}\\
Work through the problem, then give the final answer.\\[3pt]
\role{Assistant:} \lat{z_i}\\
Therefore, the final answer is:\\
Answer: \texttt{\{decoded answer\}}
\end{promptbox}

\begin{promptbox}{Final solution (after stopping; up to 512 text tokens)}
\small
\role{User:} \texttt{\{problem\}}\\
Work through the problem, then end with `Answer: \texttt{<answer>}'.\\[3pt]
\role{Assistant:} \lat{z_i}\\
Final solution (brief): \texttt{\{decoded solution, ending with ``Answer: <answer>''\}}
\end{promptbox}

In the probe and the final solution, the assistant turn begins with the agent's own belief \lat{z_i}, followed by a text cue after which the model writes text; the decoder sees only the agent's own belief, not those of the other agents. The probe is used by the stopping rule only (Algorithm~\ref{alg:beliefstream}); the final prediction is the majority over the agents' final solutions.

\clearpage
\newpage

\section{Baseline Implementation Details}
\label{app:baselines}
\vspace{10pt}
\paragraph{Multi-agent debate.}
We follow the official implementation of \citet{du2024debate} in its default configuration (three agents, two rounds, where the first round is the independent answer): three agents of the same base model receive the question with the official prompt (``Can you solve the following math problem? \ldots'' for mathematics, ``Can you answer the following question as accurately as possible? \ldots'' otherwise, with the answer requested as \texttt{\textbackslash boxed\{\}}, ``(X)'', ``(yes)/(no)'' or ``Answer: \ldots'' according to the benchmark), and in the debate round every agent is shown the full solutions of the other two agents and asked for an updated answer with the official debate prompt.

\paragraph{LatentMAS.}
We use the official implementation with the sequential-agent configuration, 40 latent steps per agent, temperature 0.6, and top-$p$ 0.95.

\begin{comment}
For non-Qwen models, we replace the Qwen-specific system prompt with a neutral prompt.
For \texttt{Gemma-3-4B}, we use the pre-normalization hidden state rescaled to the norm of the model's input embeddings.
For the tied-embedding \texttt{Llama-3.2-1B/3B}, whose output head equals the input embedding so that the implementation's least-squares realignment of the latent to the embedding space is the identity, the official latent collapses the model into incoherent text; we realign the latent through the model's own next-token distribution instead, feeding the expected input embedding $E^{\top}\mathrm{softmax}(W h)$ rescaled as in the official code, which restores coherent outputs. For \texttt{Llama-3.1-405B}, the official latent is used unchanged.
On HotpotQA, we use the implementation's \texttt{latent\_only} mode for \texttt{Gemma-3-4B} and re-feed the retained latent vectors into a fresh cache for \texttt{Phi-4-mini}.
\end{comment}

\paragraph{ThoughtComm.}
We use the official implementation with three agents and two communication rounds of 256 tokens.
For each base model, the autoencoder, communication structure, and prefix adapter are trained on 300 held-out problems following the released training schedule.
\begin{comment}
For \texttt{Phi-4-mini}, we suppress the end-of-sequence token during the first 8 decoding steps after prefix injection.
For \texttt{Llama-3.1-8B}, injected prefix vectors are rescaled to the mean token-embedding norm, with the same 8-token decoding guard.
\end{comment}

\paragraph{Interlat.}
We follow the official sender--receiver pipeline.
Sender plans are collected from 4{,}000 MATH training problems.
The receiver is trained with LoRA of rank 64 on all linear layers together with trainable input embeddings, while the hidden-state adapter is trained for 3 epochs with learning rate $10^{-4}$.
At inference time, sender hidden states are inserted between the plan delimiters and the receiver decodes greedily.

\paragraph{SoftCoT.}
We use the official implementation with 32 soft-thought tokens and train the projection on GSM8K for 3 epochs.
Decoding uses the default temperature of 1.0, with a generation budget of 4{,}096 tokens for MATH-500 and AIME and 2{,}048 tokens otherwise.

\paragraph{SwiReasoning.}
We use the official implementation with the dataset-specific hyperparameters released by the authors.
The sequential decoding budget is capped at the mean number of sequential steps used by our method on the same benchmark and base model.

\paragraph{Soft Thinking.}
We implement the method in HuggingFace Transformers following the official algorithm and defaults: at most 10 concept tokens per step, top-$p$ 0.95, top-$k$ 30, min-$p$ 0.001, temperature 0.6, and cold stopping after 256 steps with entropy below 0.01.
The sequential decoding budget is matched to our method on each benchmark and base model.
\end{document}